\documentclass[10pt,journal,compsoc]{IEEEtran}
\ifCLASSOPTIONcompsoc
  \usepackage[nocompress]{cite}
\else
  \usepackage{cite}
\fi
\usepackage{times}
\usepackage{epsfig}
\usepackage{graphicx}
\usepackage{amsmath}
\usepackage{amssymb}
\graphicspath{{figs/}}

\ifCLASSINFOpdf
\else
\fi

\usepackage[pagebackref=true,breaklinks=true,letterpaper=true,colorlinks,bookmarks=false]{hyperref}

\usepackage{algorithm}
\usepackage[noend]{algpseudocode}
\begin{document}
%
\title{Crystal Loss and Quality Pooling for Unconstrained Face Verification and Recognition}
%
%
%
%

\author{Rajeev~Ranjan,~\IEEEmembership{Member,~IEEE,}
		Ankan~Bansal,
     Hongyu~Xu,~\IEEEmembership{Member,~IEEE,}  Swami~Sankaranarayanan,~\IEEEmembership{Member,~IEEE,}      
		Jun-Cheng~Chen,~\IEEEmembership{Member,~IEEE,}
        Carlos~D~Castillo,~\IEEEmembership{Member,~IEEE,}
        and~Rama~Chellappa,~\IEEEmembership{Fellow,~IEEE}
\thanks{R. Ranjan, A. Bansal, H. Xu, S. Sankaranarayanan  and R. Chellappa are with the Department
of Electrical and Computer Engineering, University of Maryland, College Park,
MD, 20742 USA e-mail: \{rranjan1,ankan,hyxu,swamiviv,rama\}@umiacs.umd.edu.}
\thanks{J-C Chen and C. D. Castillo are with UMIACS, University of Maryland, College Park,
MD, 20742 USA e-mail: \{pullpull,carlos\}@umiacs.umd.edu.}
}

%
%

\markboth{Journal of \LaTeX\ Class Files,~Vol.~14, No.~8, August~2015}%
{Shell \MakeLowercase{\textit{et al.}}: Bare Demo of IEEEtran.cls for Computer Society Journals}
%



\IEEEtitleabstractindextext{%

\begin{abstract}
   In recent years, the performance of face verification and recognition systems based on deep convolutional neural networks (DCNNs) has significantly improved.  A typical pipeline for face verification includes training a deep network for subject classification with softmax loss, using the penultimate layer output as the feature descriptor, and generating a cosine similarity score given a pair of face images or videos. The softmax loss function does not optimize the features to have higher similarity score for positive pairs and lower similarity score for negative pairs, which leads to a performance gap. In this paper, we propose a new loss function, called Crystal Loss, that restricts the features to lie on a hypersphere of a fixed radius.  The loss can be easily implemented using existing deep learning frameworks. We show that integrating this simple step in the training pipeline significantly improves the performance of face verification and recognition systems. 
   
   Additionally, we focus on the problem of video-based face verification, where the algorithm needs to determine whether a pair of image-sets or videos belong to the same person or not. A compact feature representation is required for every video or image-set, in order to compute the similarity scores. Classical approaches tackle this problem by simply averaging the features extracted from each image/frame of the image-set/video. However, this may lead to sub-optimal feature representations since both good and poor quality faces are weighted equally. To this end, we propose Quality Pooling, which weighs the features based on input face quality. We show that face detection scores can be used as measures of face quality. We also propose Quality Attenuation, which rescales the verification score based on the face quality of a given verification pair. We achieve state-of-the-art performance for face verification and recognition on challenging LFW, IJB-A, IJB-B and IJB-C datasets over a large range of false alarm rates ($10^{-1}$ to $10^{-7}$).
   
   


\end{abstract}

}

\maketitle

\IEEEdisplaynontitleabstractindextext

%
\IEEEpeerreviewmaketitle

\IEEEraisesectionheading{\section{Introduction}\label{sec:introduction}}



\IEEEPARstart{F}{ace} verification in unconstrained settings is a challenging problem. Despite the excellent performance of recent face verification systems on datasets like Labeled Faces in the Wild (LFW)~\cite{huang2007labeled}, it is still difficult to achieve similar accuracy on faces with extreme variations in viewpoints, resolution, occlusion and image quality.  This is evident from the performance of traditional algorithms on the publicly available IJB-A~\cite{klare2015pushing} dataset. Data quality imbalance in the training set is one of the reasons for this performance gap. Existing face recognition training datasets contain large amount of high quality and frontal faces, whereas the unconstrained and difficult faces occur rarely. Most of the DCNN-based methods trained with softmax loss for classification tend to over-fit to the high quality data and fail to correctly classify faces acquired in difficult conditions.

\begin{figure}[t]
      \centering
      \includegraphics[width=7.0cm, height=4.5cm]{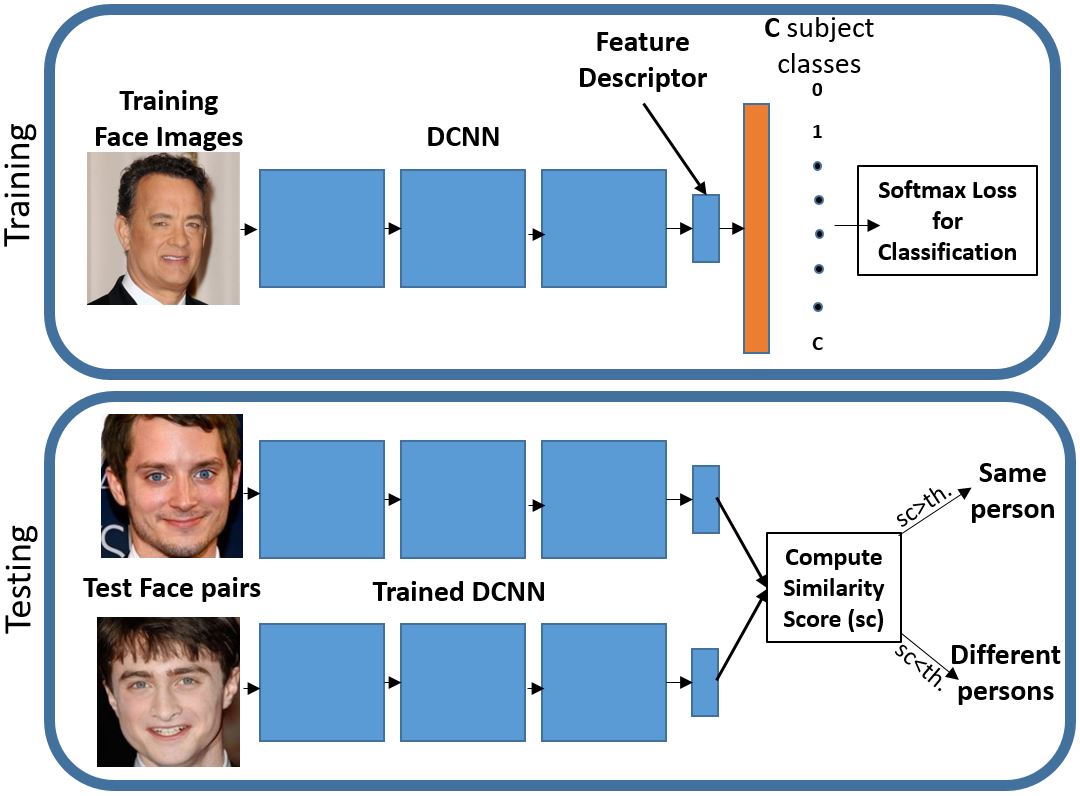}
      \caption{A general pipeline for training and testing a face verification system using DCNN.}
      \label{fig:pipeline}
\end{figure}

Using the softmax loss function for training face verification system has its own advantages and disadvantages. On the one hand, it can be easily implemented using inbuilt functions from the publicly available deep leaning toolboxes such as Caffe~\cite{jia2014caffe}, Torch~\cite{torch} and TensorFlow~\cite{tensorflow2015-whitepaper}. Unlike triplet loss~\cite{schroff2015facenet}, it does not have any restrictions on the input batch size and converges quickly. The learned features are discriminative enough for efficient face verification without any metric learning. On the other hand, the softmax loss is biased to the sample distribution. Unlike contrastive loss~\cite{sun2015deeply} and triplet loss~\cite{schroff2015facenet} which specifically attend to hard samples, the softmax loss maximizes the conditional probability of all the samples in a given mini-batch. Hence, it is suited to handle high quality faces, ignoring the rare difficult faces in a training mini-batch. We observe that the $L_{2}$-norm of features learned using softmax loss is informative of the quality of the face~\cite{parde2016deep}. Features for good quality frontal faces have a high $L_{2}$-norm while blurry faces with extreme pose have low $L_{2}$-norm (see Figure~\ref{fig:verif_norms}(b)). Moreover, the softmax loss does not optimize the verification requirement of keeping positive pairs closer and negative pairs far from each other. In order to address this limitation, many methods either apply metric learning on top of softmax features~\cite{sankaranarayanan2016triplet, chen2016unconstrained, parkhi2015deep, chen2017unconstrained} or train an auxiliary loss~\cite{wen2016discriminative, sun2015deeply, wen2016latent}  along with the softmax loss to achieve enhanced verification performance. 


In this paper, we provide a symptomatic treatment to issues associated with using softmax loss. We propose the Crystal loss function that adds a constraint on the features during training such that their $L_{2}$-norm remains constant. In other words, we restrict the features to lie on a hypersphere of a fixed radius. The proposed Crystal loss has two advantages. Firstly, it provides equal attention to both good and bad quality faces since all the features have the same $L_{2}$-norm now, which is essential for improved performance in unconstrained settings. Secondly, it strengthens the verification features by forcing the same subject features to be closer and features from different subjects to be far from each other in the normalized space. Therefore, it maximizes the margin for the normalized $L_{2}$ distance or cosine similarity score between negative and positive pairs. In this way, the proposed Crystal loss overcomes the limitations of the regular softmax loss.

The Crystal loss also retains the advantages of the regular softmax loss. Similar to the softmax loss, it is a one network, one loss system. It doesn't necessarily require any joint supervision as used by many recent methods~\cite{wen2016discriminative, parkhi2015deep, wen2016latent, sun2015deeply}. It can be easily implemented using inbuilt functions from Caffe~\cite{jia2014caffe}, Torch~\cite{torch} and TensorFlow~\cite{tensorflow2015-whitepaper}, and converges very fast. It introduces just a single scaling parameter to the network. Compared to the regular softmax loss, the Crystal loss gains a significant improvement in the performance. It achieves new state-of-the-art results on IJB-A, IJB-B, IJB-C and LFW datasets, and competitive results on YouTube Face datasets. It surpasses the performance of several state-of-the-art systems, which use multiple networks or multiple loss functions or both. Moreover, the gains from Crystal Loss are complementary to metric learning (eg: TPE~\cite{sankaranarayanan2016triplet}, joint-Bayes~\cite{chen2016unconstrained}) or auxiliary loss functions (eg: center loss~\cite{wen2016discriminative}, contrastive loss~\cite{sun2015deeply}). We show that applying these techniques on top of the Crystal Loss can further improve the verification performance. 

We also address the problem of face verification and recognition using videos or image-sets. A video may contain multiple frames with faces of a person of interest. An image-set, sometimes interchangeable with template, may contain multiple images/frames of a person of interest, captured from different sources. In a video-based or template-based face verification problem, we need to determine whether a given pair of videos/templates belong to the same identity. A traditional way to solve this problem is to represent a video or a template using a set of features, each corresponding to its constituent images or frames. This approach is not memory-efficient and does not scale with large number of videos. Additionally, computing  similarity scores between two videos for every frame-pair is of a high time complexity. Owing to these limitations, researchers have focused on generating a single feature representation from a given video or a template. 
A simple approach is to represent the video/template with arithmetic mean of the features of the constituent frames/images. This approach may lead to sub-optimal feature representation since the features for both good as well as bad quality faces get weighted equally. To this end, we propose Quality Pooling, which obtains the weight coefficients using the face detection scores. We show that these probability scores from a face detector could be treated as a measure of the face quality. A good-quality frontal face has a higher detection probability score compared to a blurry and profile face. Using the precomputed detection score does not require any additional training and improves the performance of video/template-based face verification. 

In addition, we focus on improving the face verification performance at low False Accept Rates (FARs). We propose Quality Attenuation, that rescales the similarity score based on maximum of the detection score of the verification pair. It helps in reducing the score for a dissimilar pair if the face quality of both images in the pair is poor, thus increasing the True Accept Rate~(TAR) at a given FAR. Experiments on challenging IJB-B and IJB-C datasets show that Quality Attenuation significantly improves the TARs at very low FARs.

In summary, this paper makes the following contributions:
 
\begin{enumerate}
\item We propose a simple, novel yet effective Crystal 
Loss for face verification that restricts the $L_{2}$-norm of the feature descriptor to a constant value $\alpha$.
\item We study the variations in the performance with respect to the scaling parameter $\alpha$ and provide suitable bounds on its value for achieving consistently high performance.
\item We propose Quality Pooling, which generates a compact feature representation for a video or template using face detection score.
\item We propose Quality Attenuation, which rescales the similarity scores based on the face detection scores of the verification pairs.
\item The proposed methods yields consistent and significant improvements on all the challenging face verification datasets namely LFW~\cite{huang2007labeled}, YouTube Face~\cite{liu2015targeting}, and IJB-A~\cite{klare2015pushing}, IJB-B~\cite{whitelam2017iarpa} and IJB-C~\cite{mazeiarpa}
\end{enumerate}



\section{Related Work}

In recent years, there has been significant improvements in the accuracy of face verification using deep learning methods~\cite{schroff2015facenet, taigman2014deepface, parkhi2015deep, sankaranarayanan2016triplet,Xu_XZAC_ICPR16,sun2015deeply, wen2016discriminative, ranjan2018deep}. Most of these methods have even surpassed human performance on the LFW~\cite{huang2007labeled} dataset. Although these methods use DCNNs, they differ from each other by the type of loss function used for training. For face verification, it is essential for the features of positive subject pair to be closer and features of negative subject pair to be far apart. To solve this problem, researchers have adopted two major approaches. 

In the first approach, pairs of face images are input to the training algorithm to learn a feature embedding where positive pairs are closer and negative pairs are far apart. In this direction, Chopra et al.~\cite{chopra2005learning} proposed siamese networks with contrastive loss for training. Hu et al.~\cite{hu2014discriminative} designed a  discriminative deep metric with
a margin between positive and negative face pairs. FaceNet~\cite{schroff2015facenet} introduced triplet loss to learn the metric using hard triplet face samples. 

In the second approach, the face images along with their subject labels are used to learn discriminative identification features in a classification framework. Most of the recent methods~\cite{sun2015deeply, taigman2014deepface, parkhi2015deep, yang2016neural, ranjan2018deep} train a DCNN with softmax loss to learn these features which are later used either to directly compute the similarity score for a pair of faces or to train a discriminative metric embedding~\cite{sankaranarayanan2016triplet, chen2016unconstrained}. Another strategy is to train the network for joint identification-verification task~\cite{sun2015deeply, wen2016latent, wen2016discriminative}. 
Xiong et al.~\cite{xiong2017good} proposed transferred deep feature fusion (TDFF) which involves two-stage fusion of features trained with different networks and datasets. Template adaptation~\cite{crosswhite2016template} is applied to further boost the performance.

A recent approach~\cite{wen2016discriminative} introduced center loss to learn face embeddings which have better discriminative ability. Our proposed method is different from the center loss in the following aspects. First, we use one loss function (i.e., Crystal Loss) whereas~\cite{wen2016discriminative} uses center loss jointly with the softmax loss during training. Second, center loss introduces $C \times D$ additional parameters during training where $C$ is the number of classes and $D$ is the feature dimension. On the other hand, the Crystal Loss introduces just a single parameter that defines the fixed $L_{2}$-norm of the features. Moreover, the center loss can also be used in conjunction with Crystal Loss, which performs better than center loss trained with regular softmax loss (see Section~\ref{sec:centerloss}).

Recently, some algorithms have used feature normalization during training to improve performance. Hasnat~et~al.~\cite{hasnatmises} uses class-conditional von Mises-Fisher distribution to model the feature representation. SphereFace~\cite{liu2017sphereface} proposes angular softmax (A-softmax) loss that enables DCNNs to learn angularly discriminative features. Another method called DeepVisage~\cite{hasnat2017deepvisage} uses a special case of batch normalization~\cite{ioffe2015batch} technique to normalize the feature descriptor before applying softmax loss. Our proposed method is different as it applies an $L_{2}$-constraint on the feature descriptors enforcing them to lie on a hypersphere of a given radius.

Video-based face recognition has been extensively researched in the past. Some earlier methods~\cite{lee2003video,arandjelovic2005face,turaga2011statistical,Xu_XZAC_WACV2016} represent the video frames or image-sets with appearance subspaces or manifolds. The similarity score for verification is obtained by computing manifold distances. Few other methods represent a video using local features. PEP methods~\cite{li2014eigen} cluster the local features from a part-based representation. VF$^{2}$~\cite{parkhi2014compact} aggregates Fisher Vector encodings across different video frames.

Most of the recent deep learning-based methods either use pairwise feature similarity computation for every frame-pair~\cite{taigman2014deepface,schroff2015facenet} or average frame feature pooling~\cite{chen2017unconstrained,sankaranarayanan2016triplet,ranjan2016all} to generate a video representation. The pairwise method is computation and memory expensive, while the average feature pooling treats all the features equally irrespective of the face quality. A recent method performs weighted averaging of the frame-level features, called Neural Aggregation Network (NAN)~\cite{yang2016neural}, to predict the averaging coefficients for the set of features. Similar to NAN, our proposed Quality Pooling performs weighted averaging of frame features. But, instead of predicting the coefficients, we generate them using the face detection probability score. We show that the face detection score can be used as a measure of face quality. Hence, features from good quality frames are weighed higher compared to features from poor quality frames. Thus, the proposed method generates a rich feature representation for a video without any additional expense of training a model.


\section{Motivation}

\begin{figure*}[htp!]
 \centering
\includegraphics[width=6.0cm, height=4.5cm]{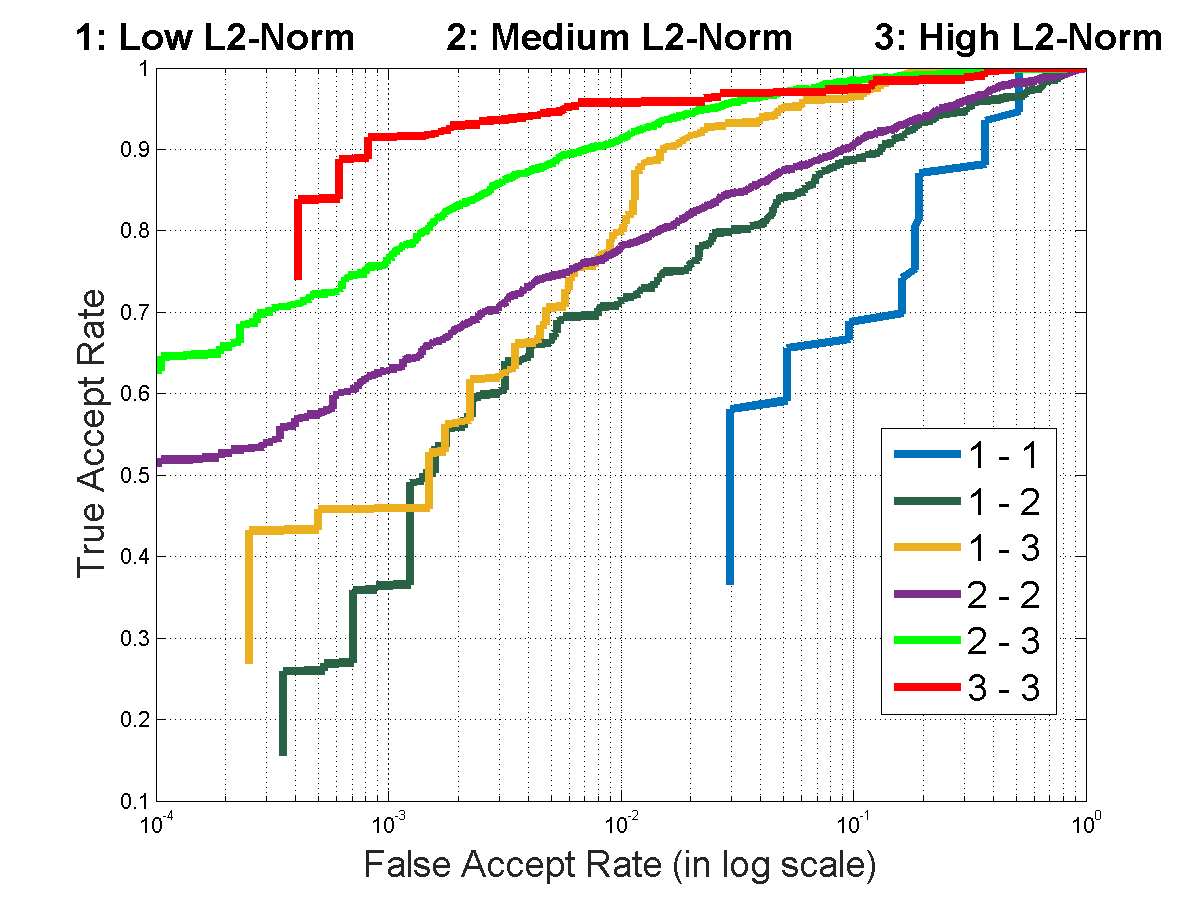}\hskip20pt\includegraphics[width=10.0cm, height=4.5cm]{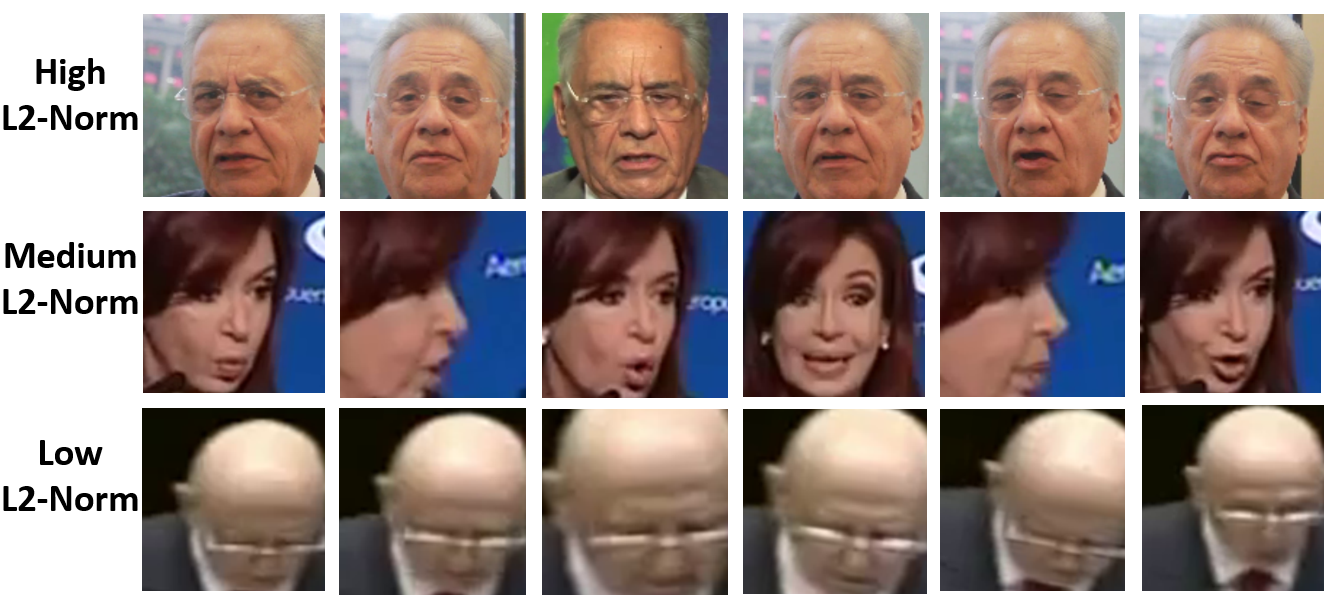}\\
(a)\hskip270pt(b)
\caption{(a) Face Verification Performance on the IJB-A dataset. The templates are divided into $3$ sets based on their $L_{2}$-norm. `1' denotes the set with low $L_{2}$-norm while `3' represents high $L_{2}$-norm. The legend `x-y' denote the evaluation pairs where one template is from set `x' and another from set `y'. (b) Sample template images from IJB-A dataset with high, medium and low L2-norm}
\label{fig:verif_norms}
\end{figure*}

We first summarize the general pipeline for training a face verification system using DCNN as shown in Figure~\ref{fig:pipeline}. Given a training dataset with face images and corresponding identity labels, a DCNN is trained as a classification task where the network learns to classify a given face image to its correct identity label. A softmax loss function is used for training the network, given by ~(\ref{eq:softmax_loss})

\begin{equation}
\label{eq:softmax_loss}
L_{S} = - \frac{1}{M} \sum_{i=1}^{M} \log \frac{e^{W_{y_{i}}^{T}f(\mathbf{x}_{i})+b_{y_{i}}}}{\sum_{j=1}^{C} e^{W_{j}^{T}f(\mathbf{x}_{i})+b_{j}}},
\end{equation}  

where $M$ is the training batch size, $\mathbf{x}_{i}$ is the $i^{th}$ input face image in the batch, $f(\mathbf{x}_{i})$ is the corresponding output of the penultimate layer of the DCNN, $y_{i}$ is the corresponding class label, and $W$ and $b$ are the weights and bias for the last layer of the network which acts as a classifier.

At test time, feature descriptors $f(\mathbf{x}_{g})$ and $f(\mathbf{x}_{p})$ are extracted for the pair of test face images $\mathbf{x}_{g}$ and $\mathbf{x}_{p}$ respectively using the trained DCNN, and normalized to unit length. Then, a similarity score is computed on the feature vectors which provides a measure of distance or how close the features lie in the embedded space. If the similarity score is greater than a threshold, the face pairs are decided to be of the same person. Usually, the similarity score is computed as the $L_{2}$-distance between the normalized features~\cite{schroff2015facenet, parkhi2015deep} or by using cosine similarity score $s$, as given by ~(\ref{eq:cosine_similarity}) ~\cite{wen2016discriminative, chen2016unconstrained, ranjan2016all, sankaranarayanan2016triplet}. Both these similarity measures are equivalent and produce same results.

\begin{equation}
\label{eq:cosine_similarity}
s = \frac{f(\mathbf{x}_{g})^{T} f(\mathbf{x}_{p})}{\|f(\mathbf{x}_{g})\|_2 \|f(\mathbf{x}_{p})\|_2}
\end{equation} 

There are two major issues with this pipeline. First, the training and testing steps for face verification task are decoupled. Training with softmax loss doesn't necessarily ensure the positive pairs to be closer and the negative pairs to be far apart in the normalized or angular space. 


Secondly, the softmax classifier is weak in modeling difficult or extreme samples. In a typical training batch with data quality imbalance, the softmax loss gets minimized by increasing the $L_{2}$-norm of the features for easy samples, and ignoring the hard samples. The network thus learns to respond to the quality of the face by the $L_{2}$-norm of its feature descriptor. To validate this claim, we perform a simple experiment on the IJB-A~\cite{klare2015pushing} dataset where we divide the templates (groups of images/frames of the same subject) into three different sets based on the $L_{2}$-norm of their feature descriptors. The features were computed using Face-Resnet~\cite{wen2016discriminative} trained with regular softmax loss. Templates with descriptors' $L_{2}$-norm \textless $90$ are assigned to set$1$. Templates with $L_{2}$-norm \textgreater $90$ but \textless $150$ are assigned to set$2$, while templates with $L_{2}$-norm \textgreater $150$ are assigned to set$3$. In total, they form six sets of evaluation pairs. Figure~\ref{fig:verif_norms}(a) shows the performance of the these six different sets for the IJB-A face verification protocol. It can be clearly seen that pairs having low $L_{2}$-norm for both templates perform very poorly, while pairs with high $L_{2}$-norm perform the best. The difference in performance between each set is quite significant. Figure~\ref{fig:verif_norms}(b) shows some sample templates from set$1$, set$2$ and set$3$ which confirms that the $L_{2}$-norm of the feature descriptor  is informative of its quality.

To solve these issues, we enforce the $L_{2}$-norm of the features to be fixed for every face image. Specifically, we add an $L_{2}$-constraint to the feature descriptor such that it lies on a hypersphere of a fixed radius. This approach has two advantages. Firstly, on a hypersphere, minimizing the softmax loss is equivalent to maximizing the cosine similarity for the positive pairs and minimizing it for the negative pairs, which strengthens the verification signal of the features. Secondly, the softmax loss is able to model the extreme and difficult faces better, since all the face features have the same $L_{2}$-norm.


\section{Proposed Method}
\label{sec:method}
The proposed Crystal Loss is given by ~(\ref{eq:l2_softmax_loss})

\begin{equation}
\label{eq:l2_softmax_loss}
\begin{aligned}
{\text{minimize}} && - \frac{1}{M} \sum_{i=1}^{M} \log \frac{e^{W_{y_{i}}^{T}f(\mathbf{x}_{i})+b_{y_{i}}}}{\sum_{j=1}^{C} e^{W_{j}^{T}f(\mathbf{x}_{i})+b_{j}}} \\
\text{subject to} && \|f(\mathbf{x}_{i})\|_2 = \alpha, ~~\forall i = 1,2,... M,\\
\end{aligned}
\end{equation}  

where $\mathbf{x}_{i}$ is the input image in a mini-batch of size $M$, $y_{i}$ is the corresponding class label, $f(\mathbf{x}_{i})$ is the feature descriptor obtained from the penultimate layer of DCNN, $C$ is the number of subject classes, and $W$ and $b$ are the weights and bias for the last layer of the network which acts as a classifier. Equation~(\ref{eq:l2_softmax_loss}) adds an additional $L_{2}$-constraint to the softmax loss defined in ~(\ref{eq:softmax_loss}). We show the effectiveness of this constraint using MNIST~\cite{lecun1998mnist} data.



\subsection{MNIST Example}
\label{sec:toy_example}

\begin{figure}[htp!]
      \centering
\includegraphics[width=3.8cm, height=3.8cm]{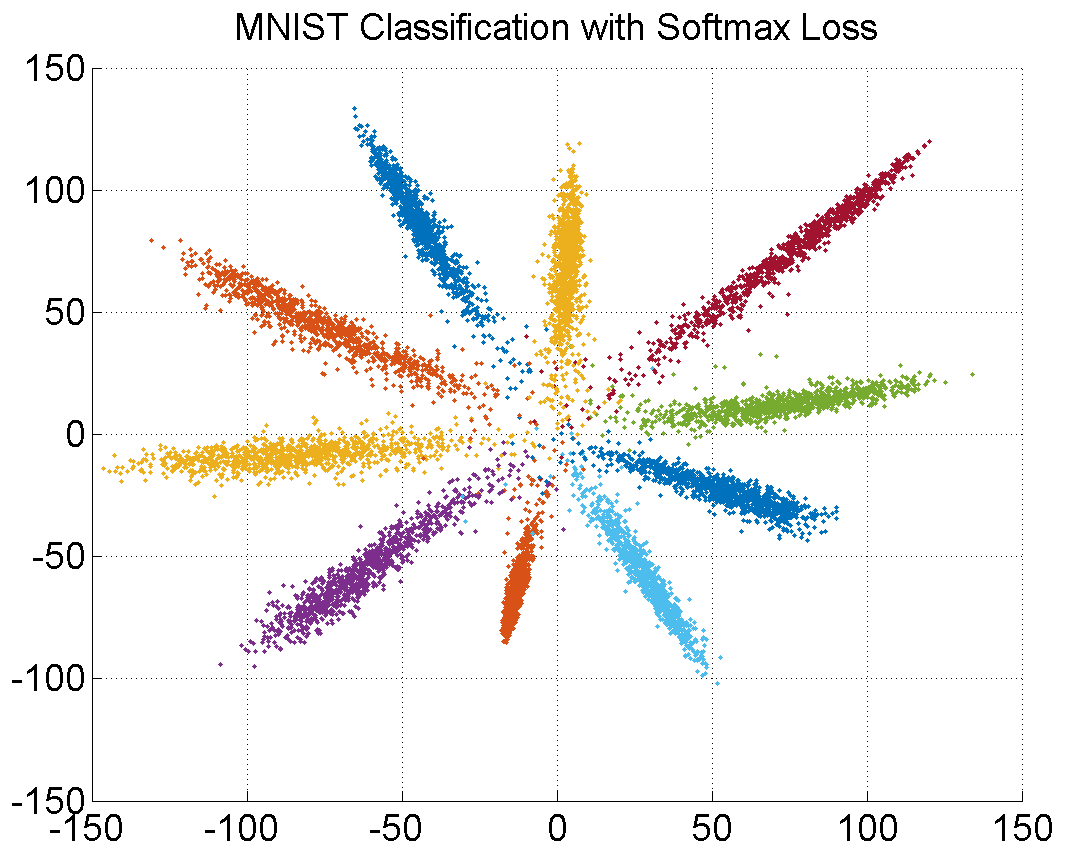}\hskip20pt\includegraphics[width=3.8cm, height=3.8cm]{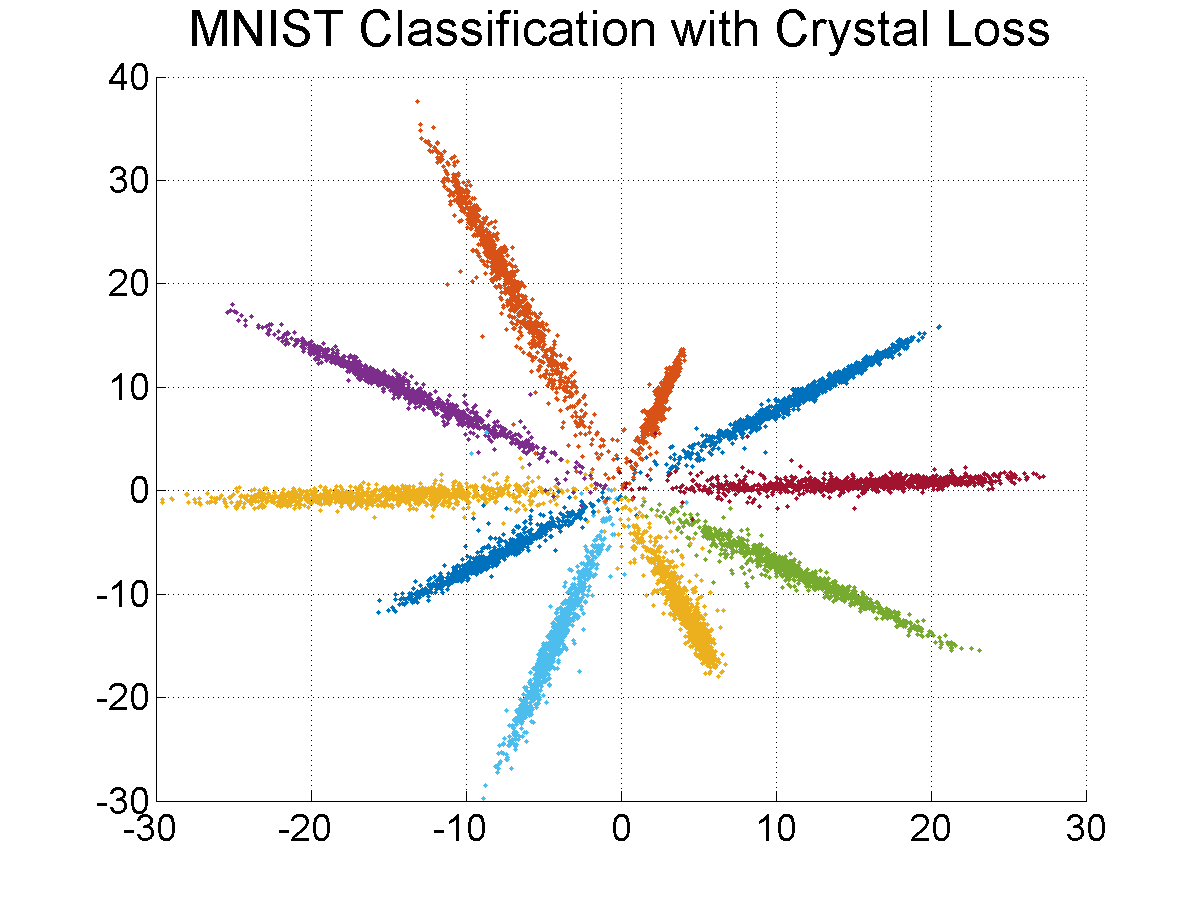}\\
(a)\hskip120pt(b)
\caption{Vizualization of $2$-dimensional features for MNIST digit classification test set using (a) Softmax Loss. (b) Crystal Loss}
      \label{fig:mnist}
\end{figure}

We study the effect of Crystal Loss on the MNIST dataset~\cite{lecun1998mnist}. We use a deeper and wider version of LeNet mentioned in~\cite{wen2016discriminative}, where the last hidden layer output is restricted to $2$-dimensional space for easy visualization. For the first setup, we train the network end-to-end using the regular softmax loss for digit classification with the number of classes equal to $10$. For the second setup, we add an $L_{2}$-normalize layer and a scale layer to the $2$-dimensional features which enforces the $L_{2}$-constraint described in~(\ref{eq:l2_softmax_loss}) (seen Section~\ref{sec:implementation} for details). Figure~\ref{fig:mnist} depicts the $2$-D features for different classes for the MNIST test set containing $10,000$ digit images. Each of the lobes shown in the figure represents $2$-D features of unique digits classes. The features for the second setup were obtained before the $L_{2}$-normalization layer.

\begin{table}[htp!]
\centering
\caption{Accuracy on MNIST test set in (\%)}
\label{tbl:mnist}
\begin{tabular}{|c|c|c|}
\hline
~ & Softmax Loss & Crystal Loss\\
\hline
Accuracy&98.88&99.05\\
\hline
\end{tabular}
\end{table}

We find two clear differences between the features learned using the two setups discussed above. First, the intra-class angular variance is large when using the regular softmax loss, which can be estimated by the average width of the lobes for each class. On the other hand, the features obtained with crystal loss have lower intra-class angular variability, and are represented by thinner lobes. Second, the magnitudes of the features are much higher with the softmax loss (ranging upto $150$), since larger feature norms result in a higher probability for a correctly classified class. In contrast, the feature norm has minimal effect on the crystal loss since every feature is normalized to a circle of fixed radius before computing the loss. Hence, the network focuses on bringing the features from the same class closer to each other and separating the features from different classes in the normalized or angular space. Table~\ref{tbl:mnist} lists the accuracy obtained with the two setups on MNIST test set. Crystal loss achieves a higher performance, reducing the error by more than $15\%$. Note that these accuracy numbers are lower compared to a typical DCNN since we are using only $2$-dimensional features for classification.

\begin{figure*}[htp!]
 \centering
\includegraphics[width=7.5cm, height=5.0cm]{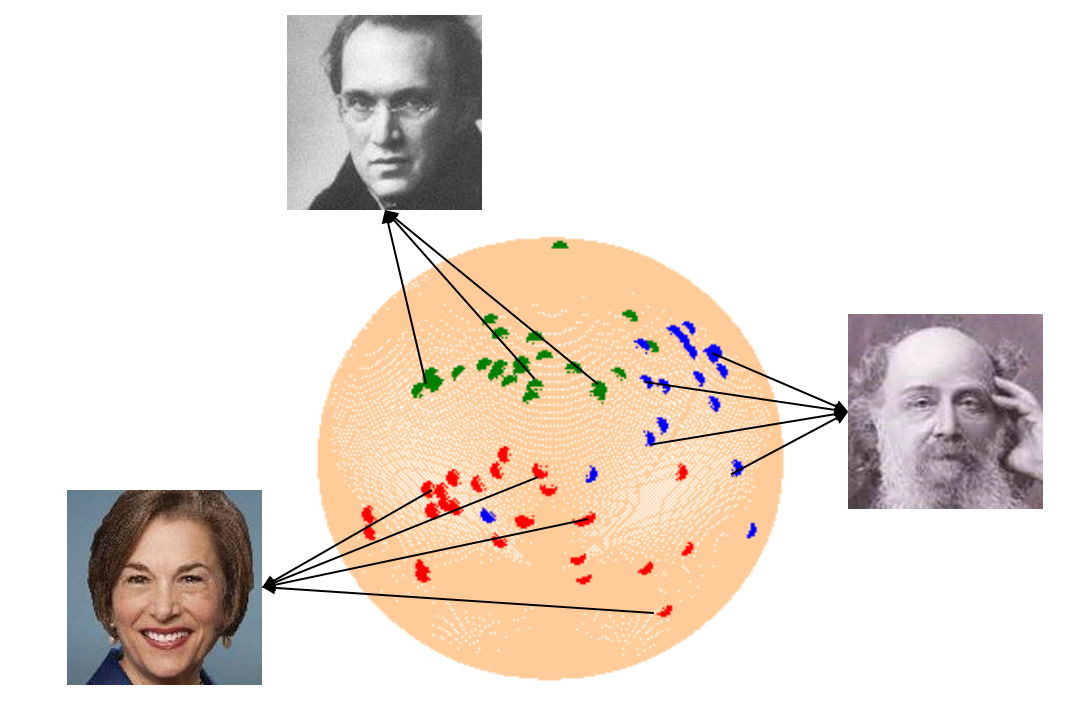}\hskip50pt\includegraphics[width=7.5cm, height=4.5cm]{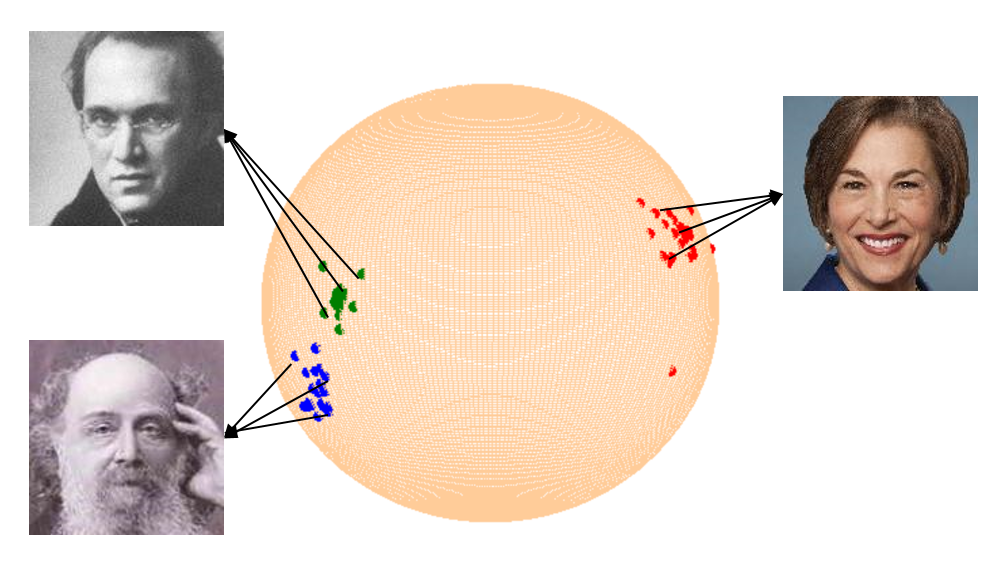}\\
(a)\hskip270pt(b)
\caption{Three-dimensional normalized features for three different identities, obtained from (a) network trained with Softmax Loss. (b) network trained with Crystal Loss. The intra-class cosine distance reduces while the inter-class cosine distance increases by using the Crystal Loss.}
\label{fig:viz_loss}
\end{figure*}

\subsection{Implementation Details}
\label{sec:implementation}
\begin{figure}[htp!]
      \centering
      \includegraphics[width=0.5\textwidth]{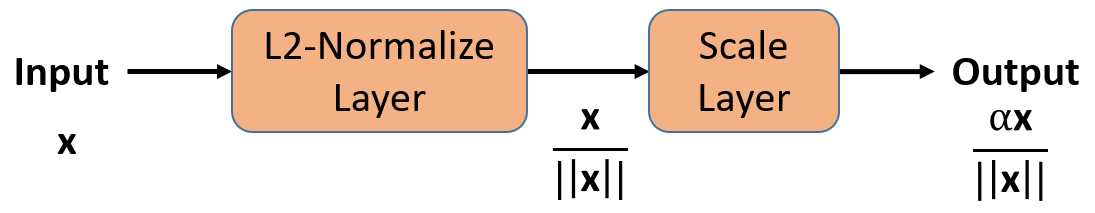}
      \caption{We add an $L_{2}$-normalization layer and a scale layer to constrain the feature descriptor to lie on a hypersphere of radius $\alpha$.}
      \label{fig:model}
\end{figure}

Here, we provide the details of implementing the $L_{2}$-constraint described in ~(\ref{eq:l2_softmax_loss}) in the framework of DCNNs. The constraint is enforced by adding an $L_{2}$-normalization layer followed by a scale layer as shown in Figure~\ref{fig:model}.

This module is added just after the penultimate layer of DCNN which acts as a feature descriptor. The $L_{2}$-normalization layer normalizes the input feature $\mathbf{x}$ to a unit vector given by~(\ref{eq:l2_normalize}). The scale layer scales the input unit vector to a fixed radius given by the parameter $\alpha$~(\ref{eq:scale}). In total, we just introduce one scalar parameter ($\alpha$) which can be trained along with the other parameters of the network.

\begin{equation}
\label{eq:l2_normalize}
\mathbf{y} = \frac{\mathbf{x}}{\|\mathbf{x}\|_2}
\end{equation}  

\begin{equation}
\label{eq:scale}
\mathbf{z} = \alpha \cdot \mathbf{y}
\end{equation}  

The module is fully differentiable and can be used in the end-to-end training of the network. At test time, the proposed module is redundant, since the features are eventually normalized to unit length while computing the cosine similarity. At training time, we backpropagate the gradients through the L2-normalization and the scale layer,  as well as compute the gradients with respect to the scaling parameter $\alpha$ using the chain rule as given below.

\begin{equation}
\begin{aligned}
\frac{\partial l}{\partial y_{i}} &= \frac{\partial l}{\partial z_{i}} \cdot \alpha \\
\frac{\partial l}{\partial \alpha} &= \sum_{j=1}^{D} \frac{\partial l}{\partial z_{j}} \cdot y_{j} \\
\frac{\partial l}{\partial x_{i}} &= \sum_{j=1}^{D} \frac{\partial l}{\partial y_{j}} \cdot \frac{\partial y_{j}}{\partial x_{i}} \\
\frac{\partial y_{i}}{\partial x_{i}} &= \frac{\|\mathbf{x}\|_2^{2} - x_{i}^{2}}{\|\mathbf{x}\|_2^{3}} \\
\frac{\partial y_{j}}{\partial x_{i}} &= \frac{- x_{i} \cdot x_{j}}{\|\mathbf{x}\|_2^{3}}
\end{aligned}
\end{equation}
\\

The features learned using Softmax Loss and Crystal Loss are shown in Figure~\ref{fig:viz_loss}. We train two networks, one with Softmax Loss and another with Crystal Loss, using $100$ training identities. We restrict the feature dimension to three for better visualization on a sphere. The blue, green and red points depict the $L_{2}$-normalized features for three different identities. It is clear from the figure that Crystal Loss forces the features to have a low intra-class angular variability and higher inter-class angular variability, which improves the face verification accuracy.

%
%
%

\subsection{Bounds on Parameter $\alpha$}
\label{sec:theory}

The scaling parameter $\alpha$ plays a crucial role in deciding the performance of $L_{2}$-softmax loss. There are two ways to enforce the $L_{2}$-constraint: 1) by keeping $\alpha$ fixed throughout the training, and 2) by letting the network to learn the parameter $\alpha$. The second way is elegant and always improves over the regular softmax loss. But, the $\alpha$ parameter learned by the network is high which results in a relaxed $L_{2}$-constraint. The softmax classifier aimed at increasing the feature norm for minimizing the overall loss, increases the $\alpha$ parameter instead, allowing it more freedom to fit to the easy samples. Hence, the $\alpha$ learned by the network forms an upper bound for the parameter. Improved performance is obtained by fixing $\alpha$ to a lower constant value.

On the other hand, with a very low value of $\alpha$, the training algorithm does not converge. For instance, $\alpha = 1$ performs poorly on the LFW~\cite{huang2007labeled} dataset, achieving an accuracy of $86.37\%$ (see Figure~\ref{fig:resnet_small}). The reason being that a hypersphere with small radius ($\alpha$) has limited surface area for embedding features from the same class together and those from different classes far from each other.

\begin{figure}[htp!]
      \centering
\includegraphics[width=3.0cm, height=3.0cm]{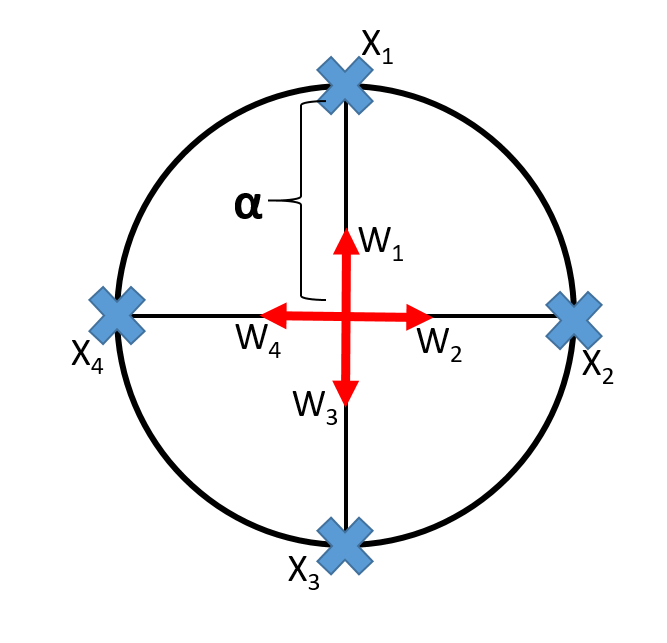}\hskip20pt\includegraphics[width=4.2cm, height=3.5cm]{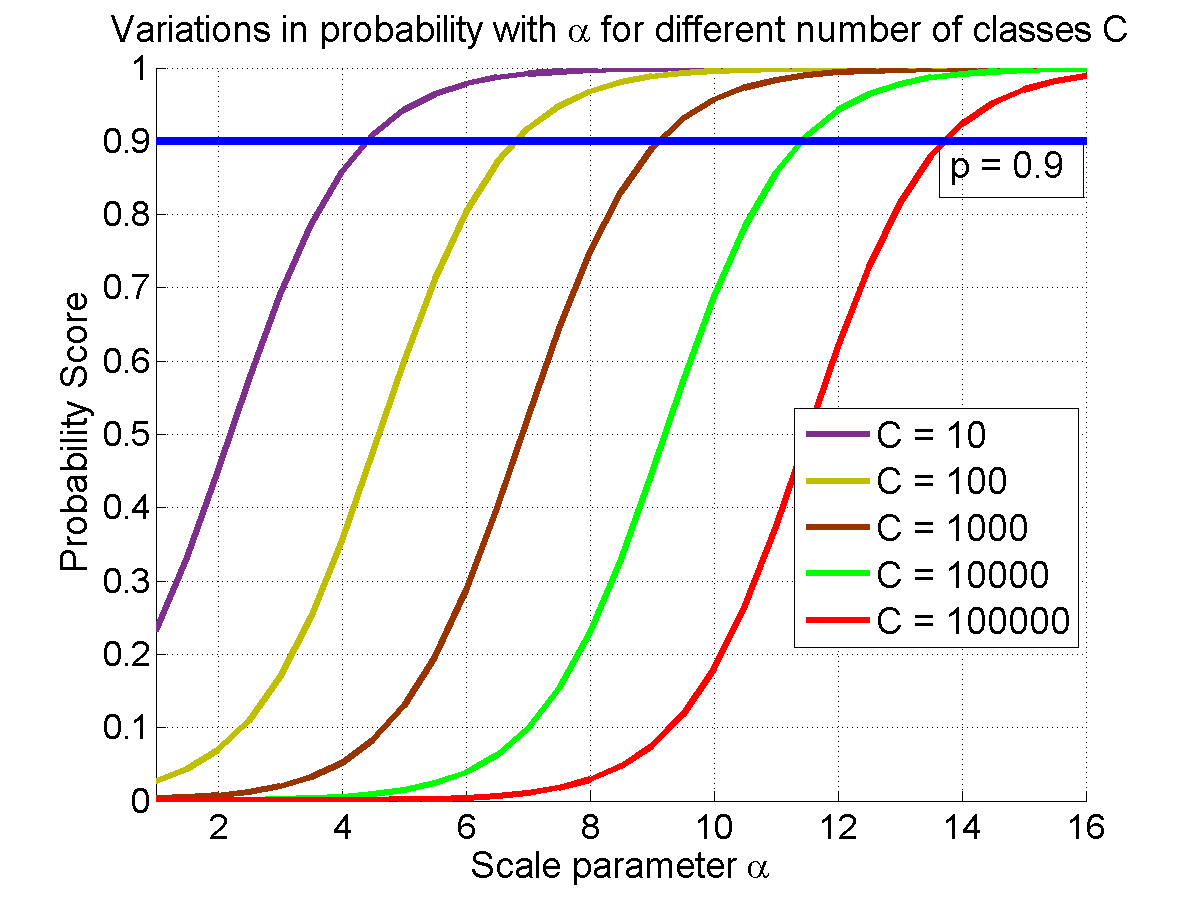}\\
(a)\hskip120pt(b)
\caption{(a) $2$-D vizualization of the assumed distribution of features (b) Variation in Softmax probability with respect to $\alpha$ for different number of classes $C$}
      \label{fig:lower_bound}
\end{figure}

Here, we formulate a theoretical lower bound on $\alpha$. Assuming the number of classes $C$ to be lower than twice the feature dimension $D$, we can distribute the classes on a hypersphere of dimension $D$ such that the centers of any two classes are at least $90^{\circ}$ apart. Figure~\ref{fig:lower_bound}(a) represents this case for $C=4$ class centers distributed on a circle of radius $\alpha$. We assume the classifier weights ($W_{i}$) to be a unit vector pointing in the direction of their respective class centers. We ignore the bias term. The average softmax probability $p$ for correctly classifying a feature is given by~(\ref{eq:2d_alpha})

\begin{equation}
\label{eq:2d_alpha}
\begin{aligned}
p &= \frac{e^{W_{i}^T X_{i}}}{\sum_{j=1}^{4} e^{W_{j}^T X_{i}}}\\
&= \frac{e^{\alpha}}{e^{\alpha} + 2 + e^{-\alpha}}
\end{aligned}
\end{equation}  

Ignoring the term $ e^{-\alpha}$ and generalizing it for $C$ classes, the average probability becomes:

\begin{equation}
\label{eq:2d_alpha1}
\begin{aligned}
p &= \frac{e^{\alpha}}{e^{\alpha} + C - 2}
\end{aligned}
\end{equation}
\\

Figure~\ref{fig:lower_bound}(b) plots the probability score as a function of the parameter $\alpha$ for various number of classes $C$. We can infer that to achieve a given classification probability (say $p = 0.9$), we need to have a higher $\alpha$ for larger $C$. Given the number of classes $C$ for a dataset, we can obtain the lower bound on $\alpha$ to achieve a probability score of $p$ by using~(\ref{eq:lower_bound}).

\begin{equation}
\label{eq:lower_bound}
\alpha_{low} = \log \frac{p (C-2)}{1-p}
\end{equation}
\\

\subsection{Relation to von Mises-Fisher Distribution}
\label{sec:vonMises}

The distribution of features learned using Crystal Loss can be characterized as a special case of von Mises-Fisher distribution~\cite{hasnatmises}. In directional statistics, von Mises-Fisher distribution is a probability distribution on a hypersphere, whose probability density function is represented using~(\ref{eq:vonMises})

\begin{equation}
\label{eq:vonMises}
f_{p}(\mathbf{x},\mathbf{\mu},\kappa) = C_{p} \exp(\kappa \mathbf{\mu}^{T} \mathbf{x}),
\end{equation}

where $\kappa \geq 0$ is the concentration parameter, $\|\mathbf{\mu}\|_{2}=1$, $\|\mathbf{x}\|_{2}=1$, and $C_{p}$ is the normalization constant dependent on $\kappa$ and the feature dimension $p$. Keeping the concentration parameter $\kappa$ same for all the $C$ classes, the log maximum a posteriori estimate for the parameters of von Mises-Fisher distribution results in the formulation of Crystal Loss ($\mathbf{L}$) as shown in~(\ref{eq:mapvmf})


\begin{equation}
\label{eq:mapvmf}
\begin{aligned}
\mathbf{L} = ~\text{maximize}~\log \frac{f_{p}(\mathbf{x}_{i},\mathbf{\mu}_{i},\kappa)}{\sum_{j=1}^{C} f_{p}(\mathbf{x}_{j},\mathbf{\mu}_{j},\kappa)}
\\
~  = \text{minimize}~ - \log \frac{\exp(\kappa \mathbf{\mu}_{i}^{T} \mathbf{x}_{i})}{\sum_{j=1}^{C} \exp(\kappa \mathbf{\mu}_{j}^{T} \mathbf{x}_{j})}\\
\end{aligned}
\end{equation}

\begin{figure}[htp!]
      \centering
\includegraphics[width=5.0cm, height=4.5cm]{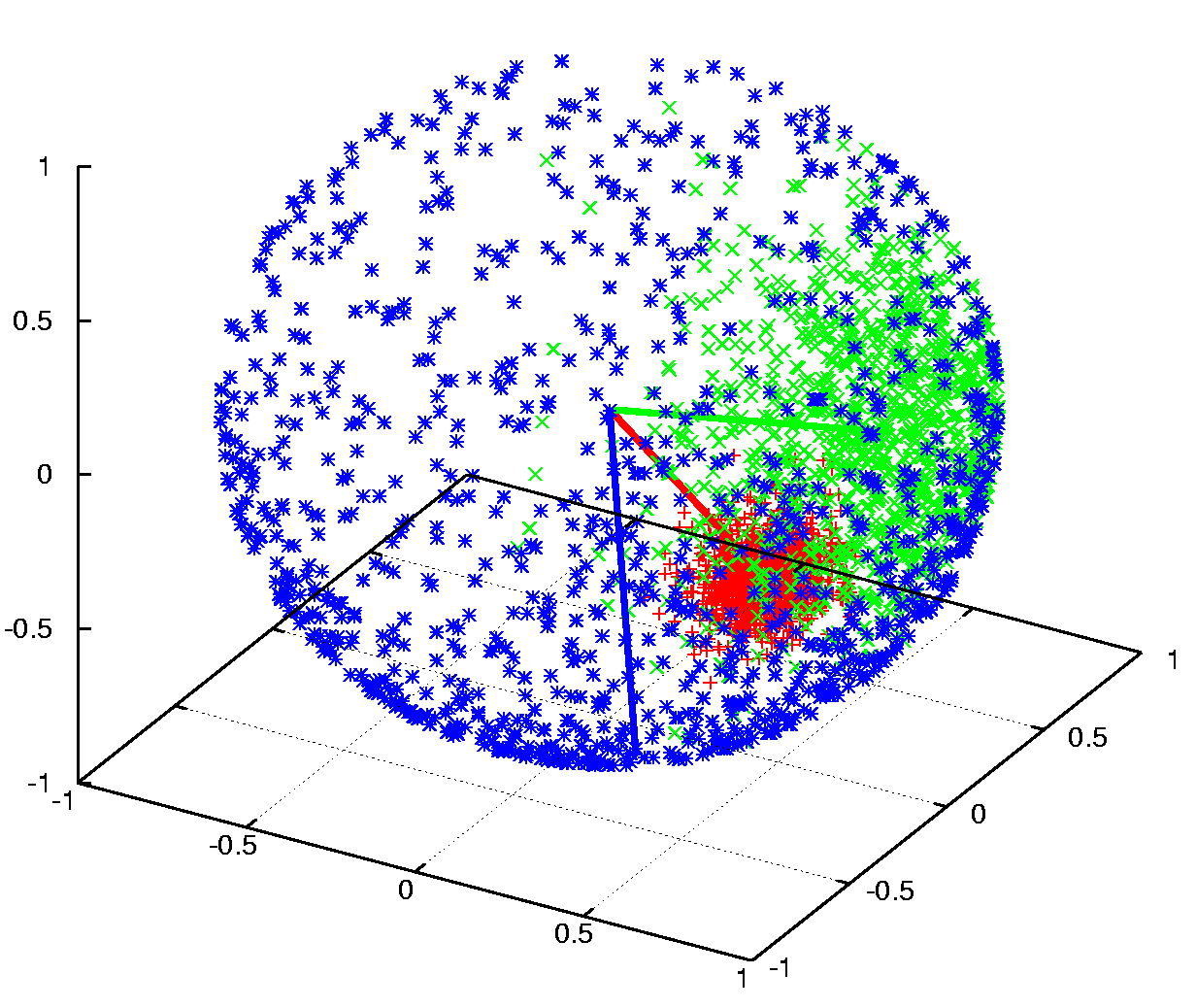}\\
\caption{Visualization of features on a sphere sampled from von Mises-Fisher distribution. The blue, green and red color represents features for different concentration parameters $\kappa = 1$, $\kappa = 10$ and $\kappa = 100$ respectively.}
      \label{fig:vmf}
\end{figure}

The concentration parameter $\kappa$ corresponds to the scale factor in the Crystal Loss. The $\kappa$ value decides the spread of the features on the hypersphere, as shown in Figure~\ref{fig:vmf}~\footnote{https://en.wikipedia.org/wiki/Von\textunderscore Mises–Fisher\textunderscore distribution}. A low value of $\kappa$ results in high intra-class angular variability, while a high value of $\kappa$ decreases the inter-class angular distance. Hence, an optimal value of $\kappa$ or the scale factor for Crystal Loss is required~(see Section~\ref{sec:theory}) so that features from same class are close together and features from different classes are far from each other in angular space. We do not normalize the classifier weight vectors since it significantly slows down the training process for large number of classes.

\section{Quality Pooling and Attenuation}
\label{sec:quality}

\begin{figure*}[htp!]
 \centering
\includegraphics[width=12.0cm, height=5.0cm]{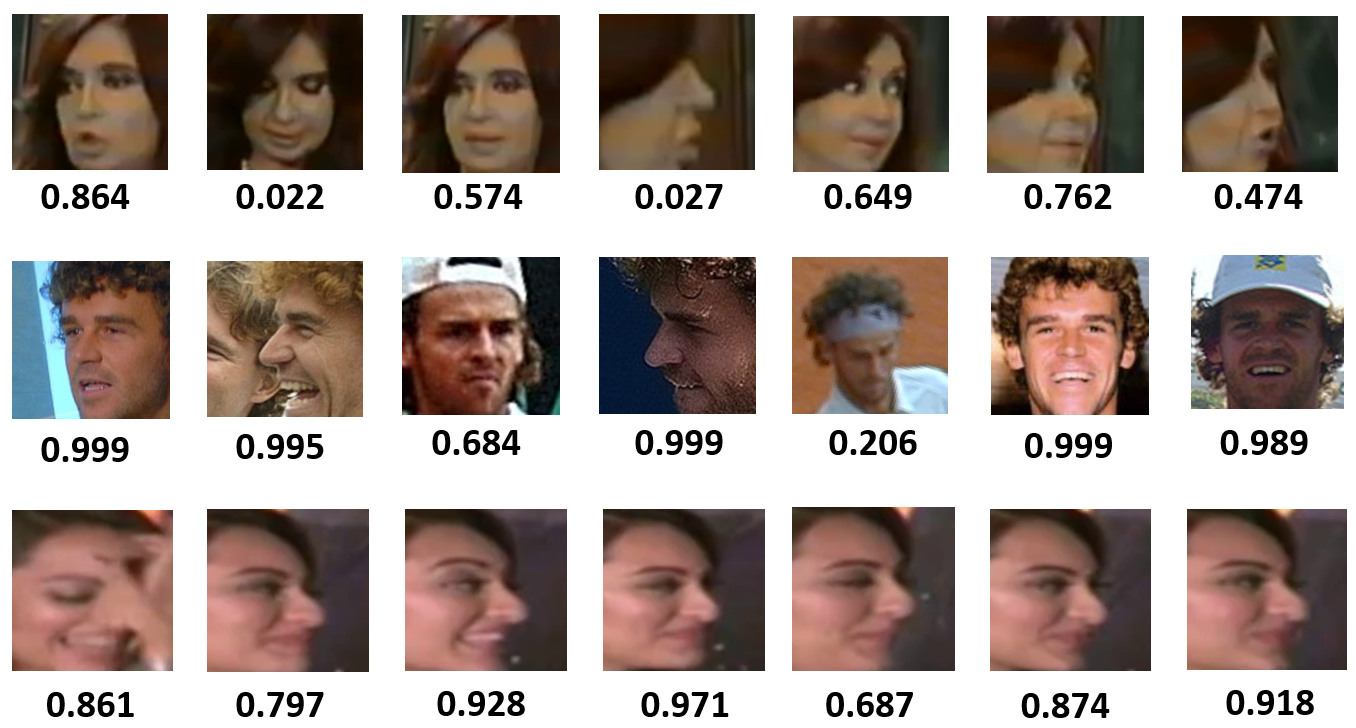}
\caption{The face detection probability score for images of three different templates (one template each row) from IJB-A dataset~\cite{klare2015pushing}. The scores reflect the face quality of the images. Higher scores correspond to good quality images, while lower scores are predicted for blurry and extreme pose faces.}
\label{fig:qp}
\end{figure*}

In this section, we propose Quality Pooling to flatten the images/frames in a template/video, and Quality Attenuation to rescale the similarity score for a verification pair. Both these methods use the precomputed face detection score obtained from a face detector. We observe that the face detection score is an indicator of the quality of a face image. A high resolution and frontal face has a higher detection probability compared to a blurry face or faces in extreme pose (see Figure~\ref{fig:qp}). We use the Single Shot Detector~\cite{liu2016ssd} trained for face detection task~\cite{ranjan2017hyperface} to generate the detection probabilities.

\subsection{Quality Pooling}
\label{sec:pooling}

Given a video/template $T$ containing set of frames/images $\{\mathbf{x}_{1},\mathbf{x}_{2},\mathbf{x}_{3},....\mathbf{x}_{k}\}$, let the corresponding feature vectors be denoted by $\{\mathbf{f}_{1},\mathbf{f}_{2},\mathbf{f}_{3},....\mathbf{f}_{k}\}$. The feature descriptor $\mathbf{r}$ for the video/template $T$ is given by~(\ref{eq:combined})

\begin{equation}
\label{eq:combined}
\mathbf{r} = \sum_{i=1}^{k} c_{i} f_{i},
\end{equation}

where $c_{i}$ is the coefficient for the weighted sum corresponding to the feature of $i^{th}$ frame/template. We can compute the coefficients as shown in~(\ref{eq:coeff})

\begin{equation}
\label{eq:coeff}
c_{i} = \frac{e^{\lambda l_{i}}}{\sum_{j=1}^{k} e^{\lambda l_{j}}},
\end{equation}

where $\lambda$ is a hyperparameter, and $l_{i}$ is the logit corresponding to the face detection probability $p_{i}$, and is given by~(\ref{eq:logit})

\begin{equation}
\label{eq:logit}
l_{i} = \min(\frac{1}{2} \log \frac{p_{i}}{1 - p_{i}}, 7).
\end{equation}

The logits are upper bounded by $7$ to avoid exponentially large values when the detection probability score is close to $1.0$. The variation in Quality Pooling performance with the hyperparameter $\lambda$ is discussed in Section~\ref{sec:ablation}. We use the value of $\lambda = 0.3$ in our experiments. Algorithm~\ref{alg:qp} provides the pseudo-code of the Quality Pooling method for generating a compact feature representation.

\begin{algorithm}
\caption{Quality Pooling}\label{alg:qp}
\begin{algorithmic}[1]
\State $k \gets number~of~frames~in~a~video$
\For{\texttt{i = 1~to~k}}
\State $ p_{i} \gets get\_detection\_score(\textbf{frame})$
\State $ \mathbf{f}_{i} \gets get\_identity\_descriptor(\textbf{frame})$
\EndFor
\State \textbf{end}
\State $\textbf{feature\_descriptor}\gets \textbf{0}$
\State $q\gets \textbf{0.3}$
\For{\texttt{i = 1~to~k}}
\State $l_{i} \gets \min(\frac{1}{2} \log \frac{p_{i}}{1 - p_{i}}, 7)$
\State $c_{i} \gets \frac{e^{\lambda l_{i}}}{\sum_{j}^{k} e^{\lambda l_{j}}}$
\State $\textbf{feature\_descriptor}\gets \textbf{feature\_descriptor} + c_{i} \mathbf{f}_{i}$
\EndFor
\State \textbf{end}
\end{algorithmic}
\end{algorithm}

\subsection{Quality Attenuation}
\label{sec:attenuation}

\begin{figure}[htp!]
      \centering
      \includegraphics[width=0.5\textwidth]{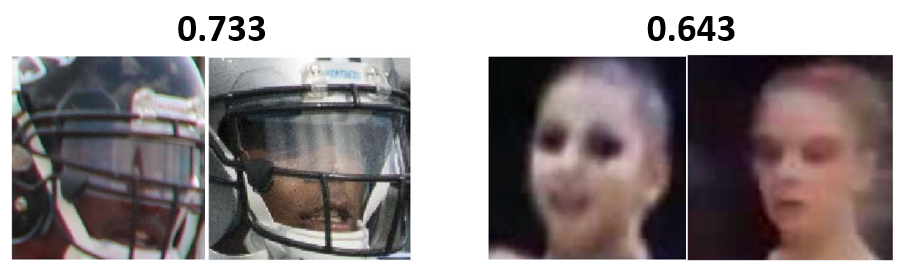}
      \caption{The non-match verification pairs from IJB-B dataset. Although the pairs have poor face quality, their cosine similarity scores (shown above the images) are high.}
      \label{fig:qa}
\end{figure}

We observe that feature descriptors for high quality faces are more discriminative, compared to those for low quality faces. Hence, the similarity scores generated for verification pairs containing both low quality faces are unreliable. This causes a non-match pair to be assigned with a high similarity score, which can significantly reduce the TAR at very low FARs of $10^{-6}$, $10^{-7}$, etc. Figure~\ref{fig:qa} shows a couple of non-match verification pairs with poor face quality, for which the network generates high similarity scores. In order to reduce the similarity scores for low quality face pairs, we scale down the similarity score by a factor of $\gamma$ if maximum face detection probability of one of the template/video in the verification pair is less than a given threshold (set to $0.75$ in our experiments). Algorithm~\ref{alg:qa} provides the pseudo-code of the proposed Quality Attenuation method.

\begin{algorithm}
\caption{Quality Attenuation}
\label{alg:qa}
\begin{algorithmic}[1]
\State $k_{1} \gets number~of~images~in~template~T{1}$
\State $k_{2} \gets number~of~images~in~template~T{2}$
\For{\texttt{i = 1~to~$k_{1}$}}
\State $ p_{i} \gets get\_detection\_score(\textbf{$T{1}_{i}$})$
\EndFor
\State \textbf{end}
\State $\textbf{lomax}_{1} \gets \max\{p_{1},p_{2},p_{3},...p_{k_{1}}\}$
\For{\texttt{i = 1~to~$k_{2}$}}
\State $ p_{i} \gets get\_detection\_score(\textbf{$T{2}_{i}$})$
\EndFor
\State \textbf{end}
\State $\textbf{lomax}_{2} \gets \max\{p_{1},p_{2},p_{3},...p_{k_{2}}\}$
\State $\textbf{score} \gets get\_similarity\_score(T{1},T{2})$
\If{$(\textbf{lomax}_{1}~or~\textbf{lomax}_{2}) \leq det\_threshold$}
\State $\textbf{score} \gets \frac{\textbf{score}}{\gamma}$
\EndIf
\State \textbf{end}
\end{algorithmic}
\end{algorithm}


\section{Results}
\label{sec:results}

\begin{figure*}[htp!]
 \centering
\includegraphics[width=18.0cm, height=3.0cm]{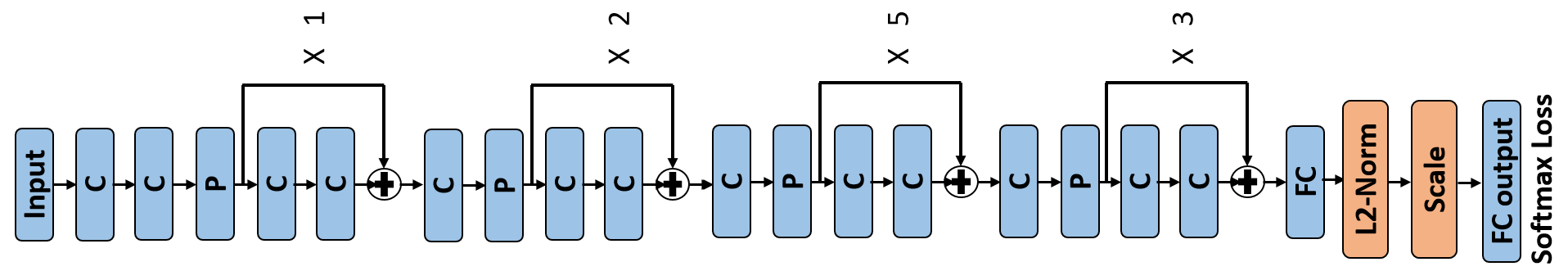}
\caption{The Face-Resnet architecture~\cite{wen2016discriminative} used for the experiments. \textbf{C} denotes Convolution Layer followed by PReLU~\cite{he2015delving} while \textbf{P} denotes Max Pooling Layer. Each pooling layer is followed by a set of residual connections, the count for which is denoted alongside. After the fully-connected layer (\textbf{FC}), we add an $L_{2}$-Normalize layer and Scale Layer which is followed by the softmax loss.}
\label{fig:architecture}
\end{figure*}

We use the publicly available Face-Resnet~\cite{wen2016discriminative} DCNN for our experiments. Figure~\ref{fig:architecture} shows the architecture of the network. It contains $27$ convolutional layers and $2$ fully-connected layers. The dimension of the feature descriptor is $512$. It utilizes the widely used residual skip-connections~\cite{he2016deep}. We add an $L_{2}$-normalization layer and a scale layer after the fully-connected layer to enforce the $L_{2}$-constraint on the descriptor. All our experiments are carried out in Caffe~\cite{jia2014caffe}.

\subsection{Baseline experiments}
In this subsection, we experimentally validate the usefulness of the $L_{2}$-softmax loss for face verification. We form two subsets of training dataset from the MS-Celeb-1M~\cite{guo2016ms} dataset: 1) MS-small containing $0.5$ million face images with the number of subjects being $13403$, and 2) MS-large containing $3.7$ million images of $58207$ subjects. The dataset was cleaned using the clustering algorithm presented in~\cite{lin2017proximity}. We train the Face-Resnet network with softmax loss as well as Crystal loss for various $\alpha$. While training with MS-small, we start with a base learning rate of $0.1$ and decrease it by ${1/10}^{th}$ after $16K$ and $24K$ iterations, upto a maximum of $28K$ iterations. For training on MS-large, we use the same learning rate but decrease it after $50K$ and $80K$ iterations upto a maximum of $100K$ iterations. A training batch size of $256$ was used. Both softmax and Crystal loss functions consume the same amount of training time which is around $9$ hours for MS-small and $32$ hours for MS-large training set respectively, on two TITAN X GPUs.  We set the learning multiplier and decay multiplier for the scale layer to $1$ for trainable $\alpha$, and $0$ for fixed $\alpha$ during the network training. We evaluate our baselines on the widely used LFW dataset~\cite{huang2007labeled} for the unrestricted setting, and the challenging IJB-A dataset~\cite{klare2015pushing} for the 1:1 face verification protocol. The faces were cropped and aligned to the size of $128 \times 128$ in both training and testing phases by implementing the face detection and alignment algorithm presented in~\cite{ranjan2016all} . 
 
\subsubsection{Experiment with small training set}
\begin{figure}[htp!]
      \centering
      \includegraphics[width=0.5\textwidth]{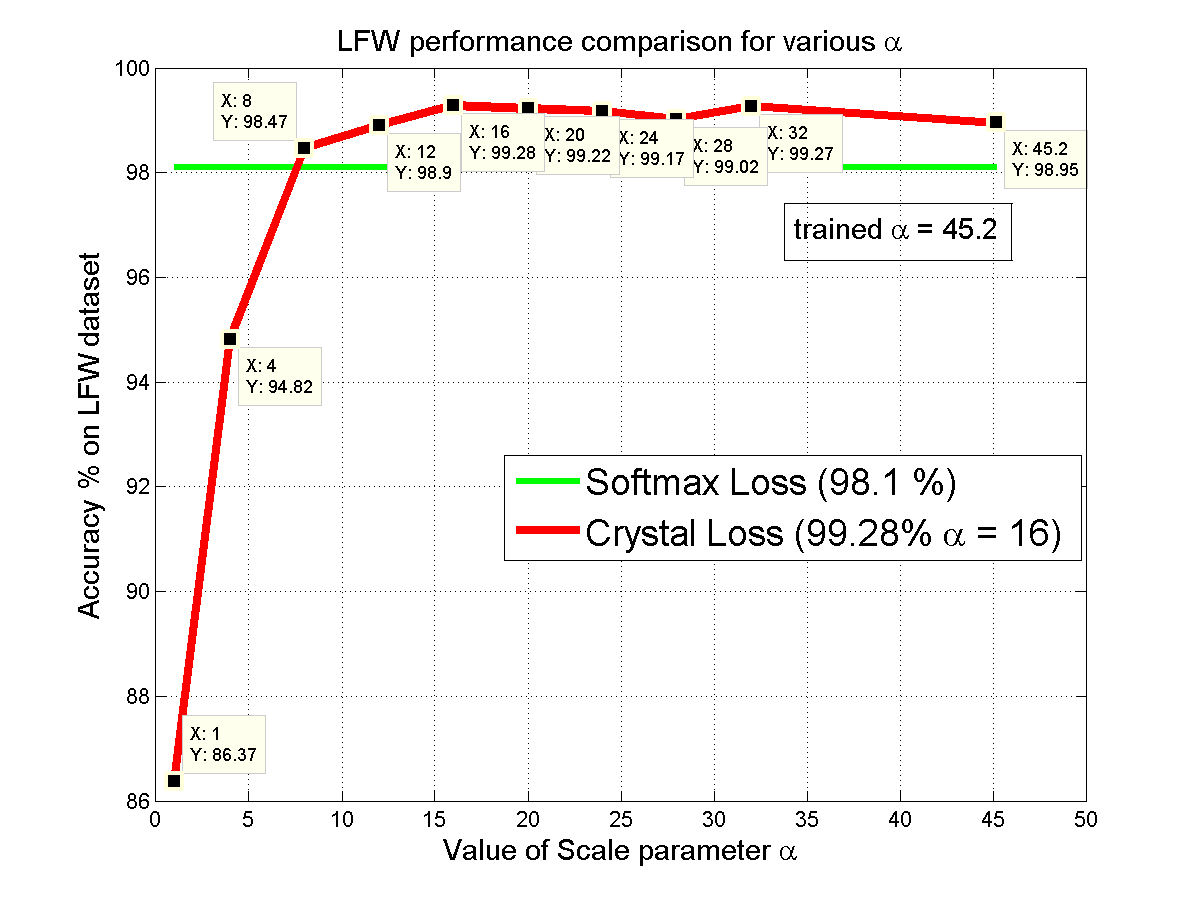}
      \caption{The red curve shows the variations in LFW accuracy with the parameter $\alpha$ for Crystal loss. The green line is the accuracy using softmax loss.}
      \label{fig:resnet_small}
\end{figure}

Here, we compare the network trained on MS-small dataset using the proposed Crystal loss, against the one trained with regular softmax loss. Figure~\ref{fig:resnet_small} shows that the softmax loss attains an accuracy of $98.1\%$ whereas the proposed Crystal loss achieves the best accuracy of $99.28\%$, thereby reducing the error by more than $62\%$. It also shows the variations in performance with the scale parameter $\alpha$. The performance is poor when $\alpha$ is below a certain threshold and stable with $\alpha$ higher than the threshold. This behavior is consistent with the theoretical analysis presented in Section~\ref{sec:theory}. From the figure, the performance of Crystal loss is better for $\alpha$ \textgreater $12$ which is close to its lower bound computed using equation~\ref{eq:lower_bound} for $C = 13403$ with a probability score of $0.9$. 

A similar trend is observed for 1:1 verification protocol on IJB-A~\cite{klare2015pushing} as shown in Table~\ref{tbl:resnet_small}, where the numbers denote True Accept Rate (TAR) at False Accept Rates (FAR) of $0.0001$, $0.001$, $0.01$ and $0.1$. Our proposed approach improves the TAR@FAR=0.0001 by $19\%$ compared to the baseline softmax loss. The performance is consistent with $\alpha$ ranging between $16$ to $32$. Another point to note is that by allowing the network to learn the scale parameter $\alpha$ by itself results in a slight decrease in performance, which shows that having a tighter constraint is a better choice.

\begin{table}[htp!]
\centering
\caption{TAR on IJB-A 1:1 Verification Protocol @FAR}
\label{tbl:resnet_small}
\tabcolsep=0.15cm
\begin{tabular}{|c|c|c|c|c|}
\hline
~ & 0.0001 & 0.001 & 0.01 & 0.1\\
\hline
Softmax Loss & 0.553 & 0.730  & 0.881 & 0.957\\
\hline
\hline
Crystal Loss ($\alpha$=8) & 0.257 & 0.433  & 0.746 & 0.953\\
\hline
Crystal Loss ($\alpha$=12) & 0.620 & 0.721  & 0.875 & 0.970\\
\hline
Crystal Loss ($\alpha$=16) & 0.734 & 0.834  & \textbf{0.924} & 0.974\\
\hline
Crystal Loss ($\alpha$=20) & 0.740 & 0.820  & 0.922 & 0.973\\
\hline
Crystal Loss ($\alpha$=24) & \textbf{0.744} & 0.831  & 0.912 & 0.974\\
\hline
Crystal Loss ($\alpha$=28) & 0.740 & \textbf{0.834}  & 0.922 & \textbf{0.975}\\
\hline
Crystal Loss ($\alpha$=32) & 0.727 & 0.831  & 0.921 & 0.972\\
\hline
Crystal Loss ($\alpha$ trained) & 0.698 & 0.817  & 0.914 & 0.971\\
\hline
\end{tabular}
\end{table}

\subsubsection{Experiment with large training set}
\begin{figure}[htp!]
      \centering
      \includegraphics[width=0.5\textwidth]{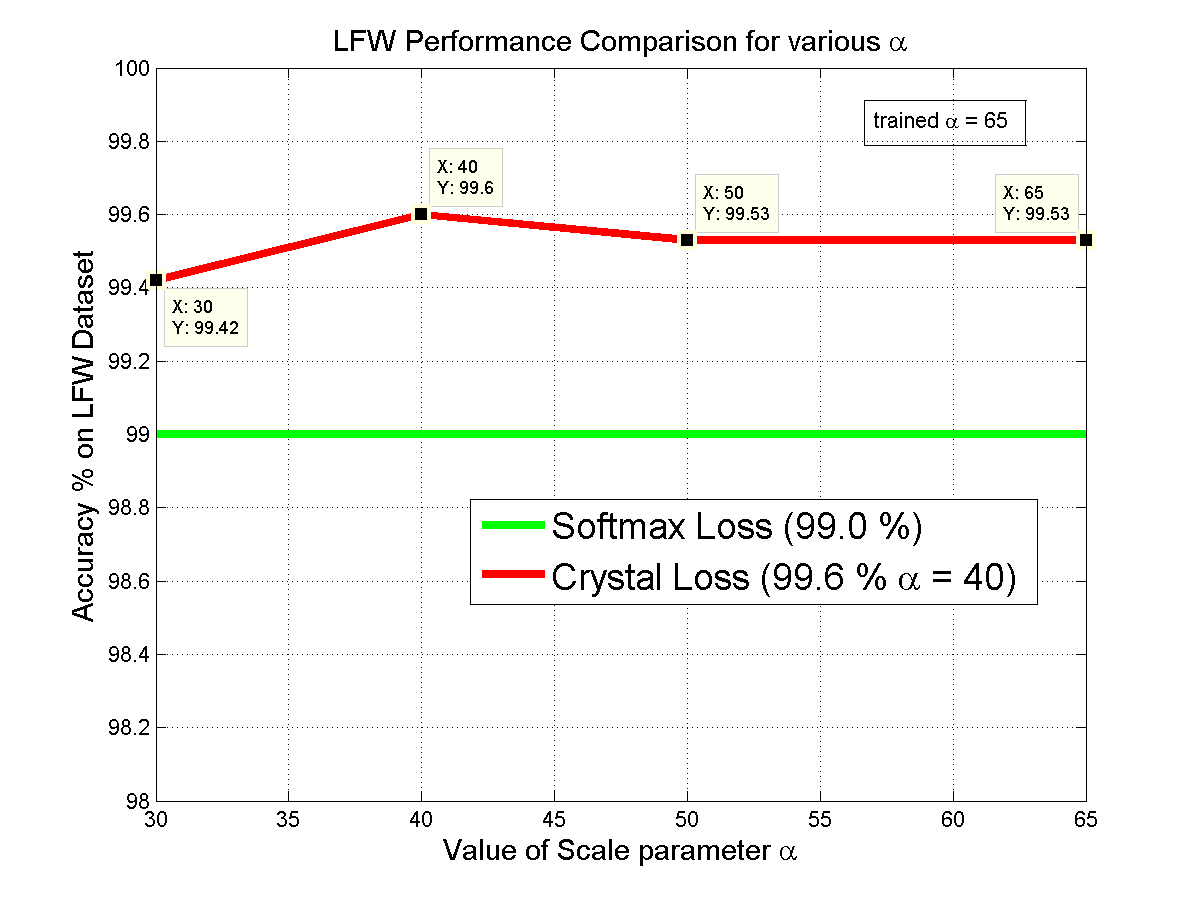}
      \caption{The red curve shows the variations in LFW accuracy with the parameter $\alpha$ for Crystal loss. The green line is the accuracy using the Softmax loss.}
      \label{fig:resnet_large}
\end{figure}

We train the network on the MS-large dataset for this experiment. Figure~\ref{fig:resnet_large} shows the performance on the LFW dataset. Similar to the small training set, the Crystal loss significantly improves over the baseline, reducing the error by $60\%$ and achieving an accuracy of $99.6\%$. Similarly, it improves the TAR@FAR=0.0001 on IJB-A by more than $10\%$ (Table~\ref{tbl:resnet_large}). The performance of Crystal loss is consistent with $\alpha$ in the range $40$ and beyond. Unlike, the small set training, the self-trained $\alpha$ performs equally good compared to fixed $\alpha$ of $40$ and $50$. The theoretical lower bound on $\alpha$ is not of much use in this case since improved performance is achieved for $\alpha$ \textgreater $30$. We can deduce that as the number of subjects increases, the lower bound on $\alpha$ is less reliable, and the self-trained $\alpha$ is more reliable with performance. This experiment clearly suggests that the proposed Crystal loss is consistent across the training and testing datasets.

\begin{table}[htp!]
\centering
\caption{TAR on IJB-A 1:1 Verification Protocol @FAR}
\label{tbl:resnet_large}
\tabcolsep=0.15cm
\begin{tabular}{|c|c|c|c|c|}
\hline
~ & 0.0001 & 0.001 & 0.01 & 0.1\\
\hline
Softmax Loss & 0.730 & 0.851  & 0.926 & 0.972\\
\hline
\hline
Crystal Loss ($\alpha$=30) & 0.775 & 0.871  & 0.938 & 0.978\\
\hline
Crystal Loss ($\alpha$=40) & 0.827 & 0.900  & 0.951 & \textbf{0.982}\\
\hline
Crystal Loss ($\alpha$=50) & \textbf{0.832} & \textbf{0.906}  & \textbf{0.952} & 0.981\\
\hline
Crystal Loss ($\alpha$ trained) & \textbf{0.832} & 0.903  & 0.950 & 0.980\\
\hline
\end{tabular}
\end{table}

\subsubsection{Experiment with a different DCNN}
\begin{figure}[htp!]
      \centering
      \includegraphics[width=0.5\textwidth]{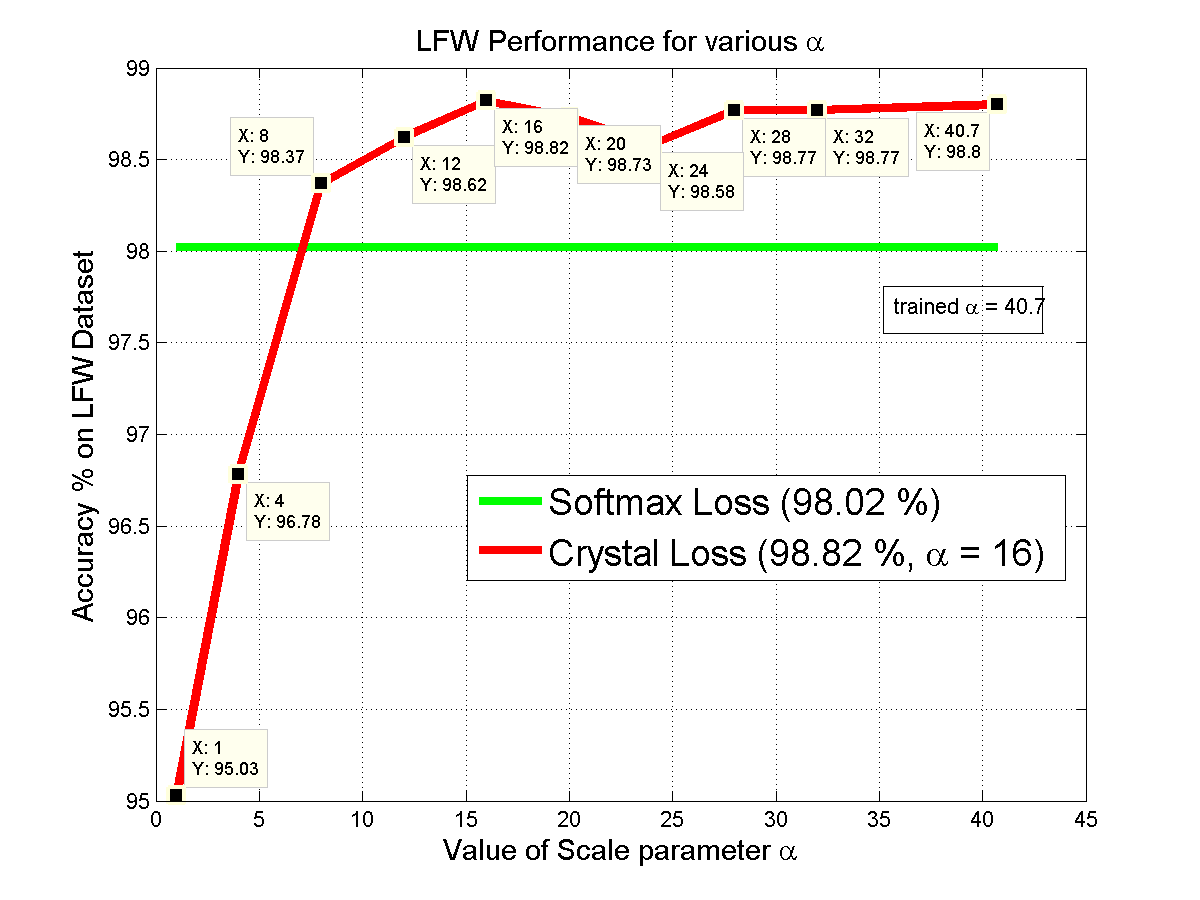}
      \caption{The red curve shows the variations in LFW accuracy with the parameter $\alpha$ for Crystal loss. The green line is the accuracy using the Softmax loss.}
      \label{fig:uf_small}
\end{figure}

To check the consistency of our proposed Crystal loss, we apply it on the All-In-One Face~\cite{ranjan2016all} instead of the Face-Resnet. We use the recognition branch of the All-In-One Face to fine-tune on the MS-small training set. The recognition branch of All-In-One Face consists of $7$ convolution layers followed by $3$ fully-connected layers and a softmax loss. We add an $L_{2}$-normalize and a scale layer after the $512$ dimension feature descriptor. Figure~\ref{fig:uf_small} shows the comparison of Crystal loss and the Softmax loss on LFW dataset. Similar to the Face-Resnet, All-In-One Face with Crystal loss improves over the Softmax performance, reducing the error by $40\%$, and achieving an accuracy of $98.82\%$. The improvement obtained by using All-In-One Face is smaller compared to the Face-Resnet. This shows that residual connections and depth of the network generate better feature embedding on a hypersphere. The performance variation with scaling parameter $\alpha$ is similar to that of Face-Resnet, indicating that the optimal scale parameter does not depend on the choice of the network. 

\begin{table}[htp!]
\centering
\caption{Accuracy on LFW (\%)}
\label{tbl:centerL2}
\begin{tabular}{|c|c|}
\hline
Softmax loss & 98.10\\
\hline
Center loss~\cite{wen2016discriminative} + Softmax loss & 99.23\\
\hline
Crystal loss & 99.28\\
\hline
Center loss~\cite{wen2016discriminative} + Crystal loss & \textbf{99.33}\\
\hline
\end{tabular}
\end{table}

\subsubsection{Experiment with auxiliary loss}
\label{sec:centerloss}

Similar to softmax loss, the Crystal loss can be coupled with auxiliary losses such as center loss, contrastive loss, triplet loss, etc. to further improve the performance. Here we study the performance variation of Crystal loss when coupled with the center loss. We use the MS-small dataset for training the networks. Table~\ref{tbl:centerL2} lists the accuracy obtained on the LFW dataset by different loss functions. The softmax loss performs the worst. The center loss improves the performance significantly when trained in conjunction with the softmax loss, and is comparable to the Crystal loss. Training center loss with Crystal loss gives the best performance of $99.33\%$ accuracy. This shows that Crystal loss is as versatile as the softmax loss and can be used efficiently with other auxiliary loss functions.

\subsection{Experiments on LFW and YTF Datasets}

\begin{table}[htp!]
\centering
\caption{Verification accuracy (in $\%$) of different methods on LFW and YTF datasets.}
\label{tbl:lfw}
\tabcolsep=0.10cm
\begin{tabular}{|c|c|c|c|c|c|}
\hline
Method & Images & $\#$nets &One loss & LFW & YTF\\
\hline
\hline
Deep Face~\cite{taigman2014deepface} & 4M & 3&No & $97.35$ & $91.4$\\
\hline
DeepID-2+~\cite{sun2015deeply} & - & 25&No & $99.47$ & $93.2$\\
\hline
FaceNet~\cite{schroff2015facenet} & 200M & 1&Yes & $99.63$ & $95.12$\\
\hline
VGG Face~\cite{parkhi2015deep} & 2.6M & 1&No & $98.95$ & $\mathbf{97.3}$\\
\hline
Baidu~\cite{liu2015targeting} & 1.3M & 1&No & $99.13$ & - \\
\hline
Wen et al.~\cite{wen2016discriminative} & 0.7M & 1&No & $99.28$ & $94.9$\\
\hline
NAN~\cite{yang2016neural} & 3M & 1&No & $-$ & $95.72$\\
\hline
DeepVisage~\cite{hasnat2017deepvisage} & 4.48M & 1&Yes & $99.62$ & $\mathbf{96.24}$\\
\hline
SphereFace~\cite{liu2017sphereface} & 0.5M & 1&Yes & $99.42$ & $95.0$\\
\hline
\hline
Softmax(FR) & 3.7M & 1&Yes & $99.0$ & $93.82$\\
\hline
CrL~(FR) & 3.7M & 1&Yes & $99.60$ & $95.54$\\
\hline
CrL~(R101) & 3.7M & 1&Yes & $99.67$ & $96.02$\\
\hline
CrL~(RX101) & 3.7M & 1&Yes & $\mathbf{99.78}$ & $\mathbf{96.08}$\\
\hline
\end{tabular}
\end{table}

We compare our algorithm with recently reported face verification methods on LFW~\cite{huang2007labeled}, YouTube Face~\cite{wolf2011face} and IJB-A~\cite{klare2015pushing} datasets. We crop and align the images for all these datasets by implementing the algorithm mentioned in~\cite{ranjan2016all}. We train the Face-Resnet (FR) with Crystal loss (CrL) as well as Softmax loss using the MS-large training set. Additionally, we train ResNet-101(R101)~\cite{he2016deep} and ResNeXt-101(RX101)~\cite{xie2016aggregated} deep networks for face recognition using MS-large training set with Crystal loss. Both R101 and RX101 models were initialized with parameters pre-trained on ImageNet~\cite{ILSVRC15} dataset. A fully-connected layer of dimension $512$ was added before the Crystal loss classifier. The scaling parameter was kept fixed with a value of $\alpha=50$. Experimental results on different datasets show that Crystal loss works efficiently with deeper models.

\begin{table*}
\caption{Face Identification and Verification Evaluation on IJB-A dataset}
\label{tbl:ijba}
\begin{center}
\tabcolsep=0.10cm
\scalebox{0.875}{
\begin{tabular}{|c|cccc||cccc|}
\hline
 & \multicolumn{4}{c|}{IJB-A Verification (TAR@FAR)} & \multicolumn{4}{c|}{IJB-A Identification}\\
\hline
Method & 0.0001 & 0.001 & 0.01 & 0.1 & FPIR=0.01 & FPIR=0.1 & Rank=1 & Rank=10\\
\hline
GOTS~\cite{klare2015pushing} & - & 0.2(0.008) & 0.41(0.014) & 0.63(0.023) & 0.047(0.02) & 0.235(0.03) & 0.443(0.02) & -\\
B-CNN~\cite{chowdhury2016one} & - & - & - & - & 0.143(0.027) & 0.341(0.032) & 0.588(0.02) & - \\
LSFS~\cite{wang2015face} & - & 0.514(0.06) & 0.733(0.034) &  0.895(0.013) & 0.383(0.063) & 0.613(0.032) & 0.820(0.024) & - \\
VGG-Face~\cite{parkhi2015deep} & - & 0.604(0.06) & 0.805(0.03) & 0.937(0.01) & 0.46(0.07) & 0.67(0.03) & 0.913(0.01) & 0.981(0.005)\\
$DCNN_{manual}$+metric~\cite{chen2015end} & - & - & 0.787(0.043) & 0.947(0.011) & - & - & 0.852(0.018) & 0.954(0.007) \\
Pose-Aware Models~\cite{masi2016pose} & - & 0.652(0.037) & 0.826(0.018) & - & -& -& 0.840(0.012) & 0.946(0.007) \\
Chen~et~al.~\cite{chen2016unconstrained} &- & - & 0.838(0.042) & 0.967(0.009) & 0.577(0.094) & 0.790(0.033) & 0.903(0.012) & 0.977(0.007)\\
Deep Multi-Pose~\cite{abdalmageed2016face} & - & - & 0.876 & 0.954 & 0.52 & 0.75 & 0.846 & 0.947 \\
Masi
et al.~\cite{masi2016we} & - & 0.725 & 0.886 & - & - & - & 0.906 & 0.977\\
Triplet Embedding~\cite{sankaranarayanan2016triplet} & - & 0.813(0.02) & 0.90(0.01) & 0.964(0.005) & 0.753(0.03) & 0.863(0.014) & 0.932(0.01) & 0.977(0.005)\\
Template Adaptation~\cite{crosswhite2016template} & - & 0.836(0.027) & 0.939(0.013) & 0.979(0.004) & 0.774(0.049) & 0.882(0.016) & 0.928(0.01) & 0.986(0.003)\\
All-In-One Face~\cite{ranjan2016all} & - & 0.823(0.02) & 0.922(0.01) & 0.976(0.004) & 0.792(0.02) & 0.887(0.014) & 0.947(0.008) & 0.988(0.003)\\
NAN~\cite{yang2016neural}  & - & 0.881(0.011) & 0.941(0.008) & 0.979(0.004) & 0.817(0.041) & 0.917(0.009) & 0.958(0.005) & 0.986(0.003)\\
FPN~\cite{chang2017faceposenet} & 0.775 & 0.852 & 0.901 & - & - & - & 0.914 & 0.938\\
TDFF~\cite{xiong2017good}  & 0.875(0.013) & 0.919(0.006) & 0.961(0.007) & 0.988(0.003) & 0.878(0.035) & 0.941(0.010) & 0.964(0.006) & \textbf{0.992(0.002)}\\
TDFF~\cite{xiong2017good}+TPE~\cite{sankaranarayanan2016triplet}  & 0.877(0.018) & 0.921(0.005) & 0.961(0.007) & \textbf{0.989(0.003)} & 0.881(0.039) & 0.940(0.009) & 0.964(0.007) & \textbf{0.992(0.003)}\\
\hline
\textbf{model-A} & \textbf{0.914(0.016)} & 0.948(0.006) & \textbf{0.971(0.004)} & 0.985(0.002) & 0.917(0.048) & \textbf{0.960(0.005)} & \textbf{0.974(0.004)} & 0.989(0.002) \\
\textbf{model-B} & 0.914(0.018) & \textbf{0.949(0.005)} & 0.969(0.003) & 0.984(0.002) & \textbf{0.918(0.043)} & 0.959(0.005) & 0.972(0.004) & 0.988(0.003) \\
\textbf{model-C} & 0.907(0.018) & 0.947(0.004) & 0.968(0.003) & 0.983(0.002) & 0.917(0.043) & 0.958(0.005) & 0.972(0.004) & 0.988(0.003) \\
\hline
\end{tabular}
}
\end{center}
\end{table*}

The LFW dataset~\cite{huang2007labeled} contains $13,233$ web-collected images from $5749$ different identities. We evaluate our model following the standard protocol of unrestricted with labeled outside data. We test on 6,000 face pairs and report the experiment results in Table~\ref{tbl:lfw}. Along with the accuracy values, we also compare with the number of images, networks and loss functions used by the methods for their overall training. The proposed method attains state-of-the-art performance with the RX101 model, achieving an accuracy of $99.78\%$. Unlike other methods which use auxiliary loss functions such as center loss and contrastive loss along with the primary softmax loss, our method uses a single loss training paradigm which makes it easier and faster to train. 

YouTube Face (YTF)~\cite{wolf2011face} dataset contains $3425$ videos of $1595$ different people, with an average length of $181.3$ frames per video. It contains $10$ folds of $500$ video pairs. We follow the standard verification protocol and report the average accuracy on splits with cross-validation in  Table~\ref{tbl:lfw}. We achieve the accuracy of $96.08\%$ using Crystal loss with RX101 network. Our method outperforms many recent algorithms and is only behind DeepVisage~\cite{hasnat2017deepvisage} which uses larger number of training samples, and VGG Face~\cite{parkhi2015deep} which further uses a discriminative metric learning on YTF.

\subsection{Experiments on IJB Datasets}

We evaluate the proposed models on three challenging IARPA Janus Benchmark datasets, namely IJB-A~\cite{klare2015pushing}, IJB-B~\cite{whitelam2017iarpa} and IJB-C~\cite{mazeiarpa}. We use Universe face dataset, a combination of curated MS-Celeb-1M~\cite{guo2016ms}, UMDFaces~\cite{bansal2016umdfaces} and UMDFaces-Videos~\cite{bansal2017s} datasets, for training the network. We remove the subject overlaps from all the three IJB-A, IJB-B and IJB-C datasets. In total, the training data contains $58,020$ subjects and $5,714,444$ images. We use ResNet-101~\cite{he2016deep} architecture with Crystal Loss for training. The scale factor $\alpha$ was set to $50$. Since the Crystal loss can be coupled with any other auxiliary loss, we use the Triplet Probabilistic Embedding (TPE)~\cite{sankaranarayanan2016triplet} to learn a $128$-dimensional embedding using the images from UMDFaces~\cite{bansal2016umdfaces} dataset. In order to showcase the effect of Quality Pooling and Quality Attenuation, we evaluate the following three models on IJB-A, IJB-B and IJB-C datasets for  the tasks of 1:1 Verification and 1:N Identification:\\

$\bullet$ \textbf{model-A} - ResNet-101 trained with Crystal Loss ($\alpha$=50), TPE~\cite{sankaranarayanan2016triplet}, media average pooling. \\

$\bullet$ \textbf{model-B} - ResNet-101 trained with Crystal Loss ($\alpha$=50), TPE~\cite{sankaranarayanan2016triplet}, Quality Pooling ($\lambda$=0.3). \\

$\bullet$ \textbf{model-C} - ResNet-101 trained with Crystal Loss ($\alpha$=50), TPE~\cite{sankaranarayanan2016triplet}, Quality Pooling ($\lambda$=0.3), Quality Attenuation ($\gamma$=1.1) \\

\subsubsection{IJB-A Dataset}

\begin{table*}
\caption{1:1 Face Verification Evaluation on IJB-B dataset}
\label{tbl:ijbb_verif}
\begin{center}
\begin{tabular}{|c|c|c|c|c|c|c|}
\hline
 & \multicolumn{6}{c|}{IJB-B 1:1 Verification (True Accept rate (in $\%$) @ False Accept Rate)}\\
\hline
Method & 0.000001 & 0.00001 & 0.0001 & 0.001 & 0.01 & 0.1 \\
\hline
\hline
GOTS~\cite{whitelam2017iarpa} & - & - & 16.0 & 33.0 & 60.0 & - \\
\hline
VGGFace~\cite{wen2016discriminative} & - & - & 55.0 & 72.0 & 86.0 & - \\
\hline
FPN~\cite{chang2017faceposenet} & - & - & 83.2 & 91.6 & 96.5 & - \\
\hline
\hline
\textbf{model-A} & 48.39 & 80.43 & 89.84 & 94.44 & \textbf{97.23} & \textbf{98.92} \\
\hline
\textbf{model-B} & \textbf{50.79} & \textbf{82.76} & \textbf{90.44} & \textbf{94.61} & 97.12 & 98.82 \\
\hline
\textbf{model-C} & 50.04 & 82.54 & 90.29 & 94.49 & 97.09 & 98.77 \\
\hline
\end{tabular}
\end{center}
\end{table*}

The IJB-A dataset~\cite{klare2015pushing} contains $500$ subjects with a total of $25,813$ images including $5,399$ still images and $20,414$ video frames. It contains faces with extreme viewpoints, resolution and illumination which makes it more challenging than the commonly used LFW dataset. The dataset is divided into $10$ splits, each containing $333$ randomly sampled subjects for training and remaining $167$ subjects for testing. Given a template containing multiple faces of the same individual, we generate a common vector representation. Table~\ref{tbl:ijba} lists the performance of recent DCNN-based methods on the IJB-A dataset. We achieve state-of-the-art result for both verification and the identification protocols. The \textbf{model-A} attains a record TAR of 0.914 @ FAR = 0.0001. Our method performs significantly better than existing methods in most of the other metrics as well. The Quality Pooling and Quality Attenuation do not show much improvement for in performance, since they are most effective at very low FARs.


\subsubsection{IJB-B Dataset}

The IJB-B dataset~\cite{whitelam2017iarpa} is an extension to the publicly available IJB-A~\cite{klare2015pushing} dataset. It contains $1,845$ unique subjects with a total of $21,798$ still images and $55,026$ video frames collected in unconstrained settings. The dataset is more challenging and diverse than IJB-A, with protocols designed to test detection, identification, verification and clustering of faces. Unlike the IJB-A dataset, it does not contain any training splits.  The verification protocol contains of $8,010,270$ between S1 and S2 gallery templates and 1:N Mixed Media probe templates. In total, they result in $10,270$ genuine comparisons and $8,000,000$ impostor comparisons. This allows us to evaluate the performance at very low FARs of $10^{-5}$ and $10^{-6}$. We compare our proposed  methods with Government-off-the-shelf (GOTS~\cite{whitelam2017iarpa}), VGGFace~\cite{wen2016discriminative} and FacePoseNet (FPN~\cite{chang2017faceposenet}). Table~\ref{tbl:ijbb_verif} lists the performance of various methods on 1:1 Verification protocol of IJB-B dataset. We achieve significant improvement over the previous state-of-the-art methods, with TAR of $50.79\%$ at FAR = $10^{-6}$. Table~\ref{tbl:ijbb_iden} provides the results for 1:N identification protocol on IJB-B dataset. We evaluate both open-set and closed-set protocol. Since, the dataset contains two set of galleries S1 and S2, we report the average performance of both the gallery sets. We achieve a True Positive Identification Rate (TPIR) of $43.44\%$ at False Positive Identification Rate (FPIR) of $0.1\%$ in the open-set protocol, and a Rank-1 accuracy of $93.69\%$ in the closed-set protocol. The proposed methods perform significantly better than previous state-of-the-art algorithms. The results show that Quality Pooling performs better than the naive media averaging of templates. We do not see much improvement with Quality Attenuation, since its more applicable to very low FARs and FPIRs.

\begin{table}
\caption{1:N Face Identification Evaluation on IJB-B dataset}
\label{tbl:ijbb_iden}
\begin{center}
\begin{tabular}{|c|c|c|c|c|c|}
\hline
 & \multicolumn{5}{c|}{IJB-B 1:N Identification}\\
\hline
& \multicolumn{3}{c|}{TPIR ($\%$) @ FPIR} & \multicolumn{2}{c|}{Retrieval Rate ($\%$)} \\
\hline
Method & 0.001 & 0.01 & 0.1 & Rank=1 & Rank=10 \\
\hline
\hline
GOTS~\cite{whitelam2017iarpa} & - & - & - & 42.0 & 62.0 \\
\hline
VGGFace~\cite{wen2016discriminative} & - & - & - & 78.0 & 89.0 \\
\hline
FPN~\cite{chang2017faceposenet} & - & - & - & 91.1 & 96.5 \\
\hline
\hline
\textbf{model-A} & \textbf{43.44} & 82.75 & 91.55 & 93.56 & \textbf{97.28} \\
\hline
\textbf{model-B} & 39.55 & \textbf{83.59} & \textbf{91.91} & \textbf{93.69} & 97.19 \\
\hline
\textbf{model-C} & 38.92 & 82.96 & 91.67 & 93.59 & 97.18 \\
\hline
\end{tabular}
\end{center}
\end{table}

\begin{table*}
\caption{Face Verification Evaluation on IJB-C dataset}
\label{tbl:ijbc_verif}
\begin{center}
\begin{tabular}{|c|c|c|c|c|c|c|c|}
\hline
 & \multicolumn{7}{c|}{IJB-C 1:1 Verification (True Accept rate (in $\%$) @ False Accept Rate)}\\
\hline
Method & 0.0000001 & 0.000001 & 0.00001 & 0.0001 & 0.001 & 0.01 & 0.1 \\
\hline
\hline
GOTS~\cite{mazeiarpa} & - & 3.00 & 6.61 & 14.67 & 33.04 & 61.99 & 80.93 \\
\hline
FaceNet~\cite{schroff2015facenet} & - & 20.95 & 33.30 & 48.69 & 66.45 & 81.76 & 92.45 \\
\hline
VGGFace~\cite{parkhi2015deep} & - & 32.20 & 43.69 & 59.75 & 74.79 & 87.13 & 95.64 \\
\hline
\hline
\textbf{model-A} & 65.96 & 76.46 & 86.25 & 91.91 & 95.72 & \textbf{97.83} & \textbf{99.14} \\
\hline
\textbf{model-B} & 66.55 & 77.91 & \textbf{87.75} & 92.50 & \textbf{95.78} & 97.75 & 99.09 \\
\hline
\textbf{model-C} & \textbf{71.37} & \textbf{81.15} & 87.35 & 92.29 & 95.63 & 97.66 & 99.06 \\
\hline
\hline
Janus1 & - & 62.93 & 76.37 & 87.13 & 93.90 & 97.43 & 99.01 \\
\hline
Janus2 & - & 49.95 & 76.21 & 85.99 & 92.31 & 95.73 & 97.15 \\
\hline
\textbf{UMD} & - & 80.82 & 87.62 & \textbf{92.52} & 95.75 & 97.63 & 99.03 \\
\hline
\end{tabular}
\end{center}
\end{table*}

\subsubsection{IJB-C Dataset}

The IJB-C dataset~\cite{mazeiarpa} is an extension to the publicly available IJB-B~\cite{whitelam2017iarpa} dataset. It contains $3,531$ unique subjects with a total of $31,334$ still images and $117,542$ video frames collected in unconstrained settings. Similar to IJB-B dataset, the protocols are designed to test detection, identification, verification and clustering of faces. The dataset also contains end-to-end protocols to evaluate the algorithm's ability to perform end-to-end face recognition. The verification protocol contains $19,557$ genuine comparisons and $15,638,932$ impostor comparisons. This allows us to evaluate the performance at very low FARs of $10^{-6}$ and $10^{-7}$. 

We compare our proposed methods with Government-off-the-shelf (GOTS~\cite{mazeiarpa}), VGGFace~\cite{parkhi2015deep} and FaceNet~\cite{schroff2015facenet}. 
We also report the performance of three Janus systems, namely Janus1, Janus2 and UMD, on IJB-C~\cite{mazeiarpa} dataset. The results of Janus performers were obtained in a private communication as of January 16, 2018. The UMD system performs score-level fusion of \textbf{model-C} with three other networks: 1) ResNet-101~\cite{he2016deep} trained on MSCeleb-1M dataset with Crystal Loss, 2) Inception-ResNet-V2~\cite{szegedy2017inception} trained on Universe face dataset with Softmax Loss, and 3) All-In-One Face~\cite{ranjan2016all} trained on MSCeleb-1M dataset using Softmax Loss. Quality Pooling and Quality Attenuation is applied to all the networks involved in fusion. The score-level fusion from different networks improves the performance in most of the metrics.

Table~\ref{tbl:ijbc_verif} lists the performance of the proposed methods on 1:1 Verification protocol of IJB-C dataset. We achieve state-of-the-art results with TARs of $71.37\%$ and $81.15\%$ at FARs = $10^{-7}$ and $10^{-6}$ respectively. Table~\ref{tbl:ijbc_iden} provides the results for 1:N identification protocol on IJB-C dataset, where the average performance on gallery sets G1 and G2 are reported. We achieve a TPIR of $78.54\%$ at FPIR of $0.1\%$ in the open-set protocol, and a Rank-1 accuracy of $94.73\%$ in the closed-set protocol. The proposed methods outperform previously state-of-the-art algorithms by a large margin. Quality Pooling shows improvement for most of the metrics, whereas Quality Attenuation improves low FARs of $10^{-6}$ and $10^{-7}$ for the verification protocol and low FPIR of $0.1\%$ for the identification protocol.

\begin{table}
\caption{1:N Face Identification Evaluation on IJB-C dataset}
\label{tbl:ijbc_iden}
\begin{center}
\begin{tabular}{|c|c|c|c|c|c|}
\hline
 & \multicolumn{5}{c|}{IJB-C 1:N Identification}\\
\hline
& \multicolumn{3}{c|}{TPIR ($\%$) @ FPIR} & \multicolumn{2}{c|}{Retrieval Rate ($\%$)} \\
\hline
Method & 0.001 & 0.01 & 0.1 & Rank=1 & Rank=10 \\
\hline
\hline
GOTS~\cite{mazeiarpa} & 2.66 & 5.78 & 15.60 & 37.85 & 60.24 \\
\hline
FaceNet~\cite{schroff2015facenet} & 20.58 & 32.40 & 50.98 & 69.22 & 81.36 \\
\hline
VGGFace~\cite{parkhi2015deep} & 26.18 & 45.06 & 62.75 & 78.60 & 89.20 \\
\hline
\hline
\textbf{model-A} & 78.42 & 86.09 & 91.91 & 94.56 & 97.53 \\
\hline
\textbf{model-B} & 78.16 & 87.38 & 92.27 & 94.64 & 97.51 \\
\hline
\textbf{model-C} & \textbf{78.54} & 87.01 & 92.10 & 94.57 & 97.48 \\
\hline
\hline
Janus1 & 69.54 & 81.67 & 89.83 & 94.03 & \textbf{97.75} \\
\hline
Janus2 & 36.03 & 75.74 & 85.70 & 89.77 & 94.95 \\
\hline
\textbf{UMD} & 78.28 & \textbf{87.58} & \textbf{92.38} & \textbf{94.73} & 97.44 \\
\hline
\end{tabular}
\end{center}
\end{table}

\subsection{Effect of Quality Pooling and Quality Attenuation}
\label{sec:ablation}

Here, we provide the ablation study of using the Quality Pooling and Quality Attenuation modules for  IJB-C~\cite{mazeiarpa} 1:1 verification protocol. 
Table~\ref{tbl:qp} shows the variations in performance for different $\lambda$ parameter setting of Quality Pooling. $\lambda = 0$ corresponds to the simple media averaging technique used in~\cite{sankaranarayanan2016triplet}. We observe that the performance improves by increasing $\lambda$ upto a value of $0.2$, after which it decreases consistently for all FARs. This proves that the DCNN feature descriptors are more reliable for good quality faces, which makes Quality Pooling perform better than naive media averaging.

Table~\ref{tbl:qa} shows the variations in performance for different Quality Attenuation parameter $\gamma$. A $\gamma$ value of $1.0$ corresponds to no attenuation in similarity score. From the table, we see that Quality Attenuation with $\gamma = 1.1$ significantly improves the performance at very low FARs of $10^{-7}$ and $10^{-6}$ with negligible decrease in performance at high FARs. Thus, Quality Attenuation is quite helpful in the face recognition system where false positives are highly undesirable.

\begin{table}
\caption{Effect of Quality Pooling on IJB-C 1:1 Verification}
\label{tbl:qp}
\begin{center}
\begin{tabular}{|c|c|c|c|c|c|c|}
\hline
 & \multicolumn{6}{c|}{Quality Pooling parameter $\lambda$}\\
\hline
TAR@FAR & 0 & 0.1 & 0.2 & 0.3 & 0.4 & 0.5 \\
\hline
\hline
$10^{-7}$ & 65.96 & 67.54 & \textbf{67.98} & 66.55 & 64.12 & 61.73 \\
\hline
$10^{-6}$ & 76.46 & 78.68 & \textbf{78.95} & 77.91 & 77.28 & 76.26 \\
\hline
$10^{-5}$ & 86.25 & 87.61 & \textbf{88.00} & 87.75 & 87.36 & 86.95 \\
\hline
$10^{-4}$ & 91.91 & 92.53 & \textbf{92.54} & 92.50 & 92.16 & 91.82 \\
\hline
$10^{-3}$ & 95.72 & 95.83 & \textbf{95.85} & 95.78 & 95.55 & 95.36 \\
\hline
$10^{-2}$ & 97.83 & \textbf{97.90} & 97.83 & 97.75 & 97.64 & 97.54 \\
\hline
$10^{-1}$ & 99.14 & \textbf{99.15} & 99.13 & 99.09 & 99.04 & 99.00 \\
\hline
\end{tabular}
\end{center}
\end{table}

\begin{table}
\caption{Effect of Quality Attenuation on IJB-C 1:1 Verification}
\label{tbl:qa}
\begin{center}
\begin{tabular}{|c|c|c|c|c|c|}
\hline
 & \multicolumn{5}{c|}{Quality Attenuation parameter $\gamma$}\\
\hline
TAR@FAR & 1.0 & 1.1 & 1.2 & 1.3 & 1.4 \\
\hline
\hline
$10^{-7}$ & 66.55 & \textbf{71.37} & 70.68 & 70.47 & 70.46  \\
\hline
$10^{-6}$ & 77.91 & \textbf{81.15} & 81.03 & 80.39 & 80.07 \\
\hline
$10^{-5}$ & \textbf{87.75} & 87.35 & 86.70 & 86.17 & 85.59 \\
\hline
$10^{-4}$ & \textbf{92.50} & 92.29 & 91.99 & 91.48 & 90.84 \\
\hline
$10^{-3}$ & \textbf{95.78} & 95.63 & 95.45 & 95.21 & 94.99 \\
\hline
$10^{-2}$ & \textbf{97.75} & 97.66 & 97.57 & 97.48 & 97.37 \\
\hline
$10^{-1}$ & \textbf{99.09} & 99.06 & 99.02 & 98.99 & 98.95 \\
\hline
\end{tabular}
\end{center}
\end{table}

\section{Discussion}
\label{sec:discuss}

\begin{figure}[htp!]
      \centering
\includegraphics[width=8.0cm, height=3.0cm]{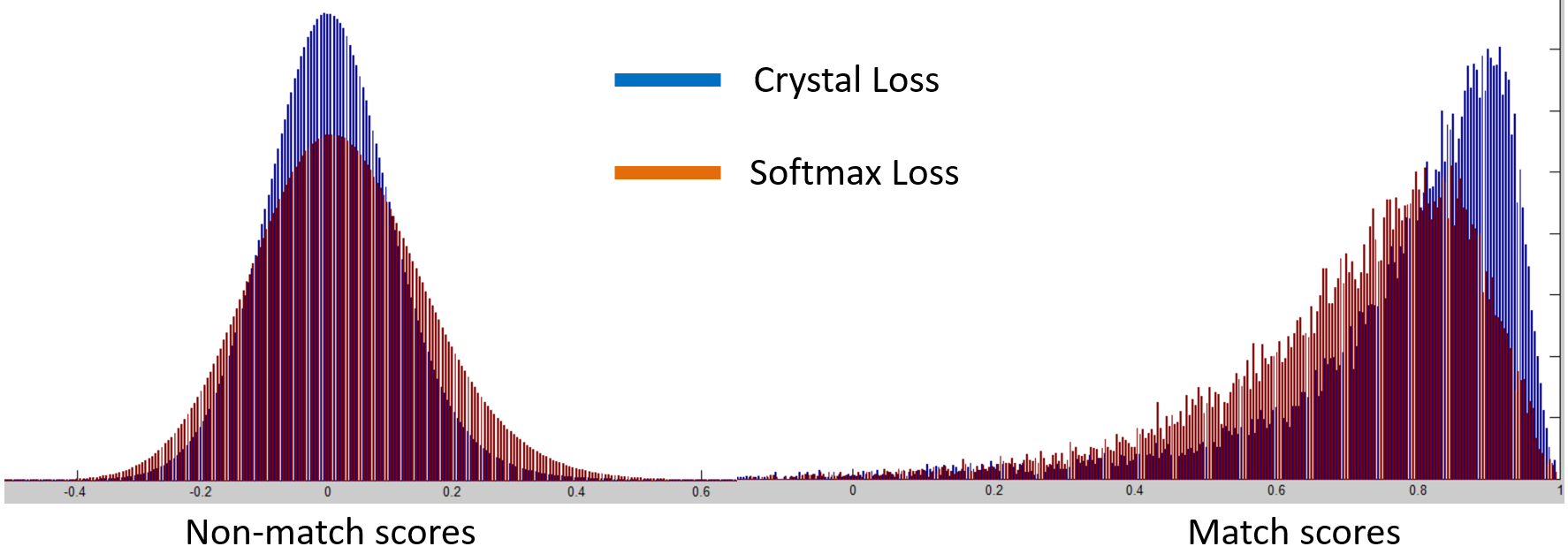}\\
\caption{Similarity Score distribution with Crystal Loss and Softmax Loss for face verification pairs of IJB-C~\cite{mazeiarpa} dataset.}
      \label{fig:score}
\end{figure}

\begin{figure*}[htp!]
 \centering
\includegraphics[width=1\linewidth]{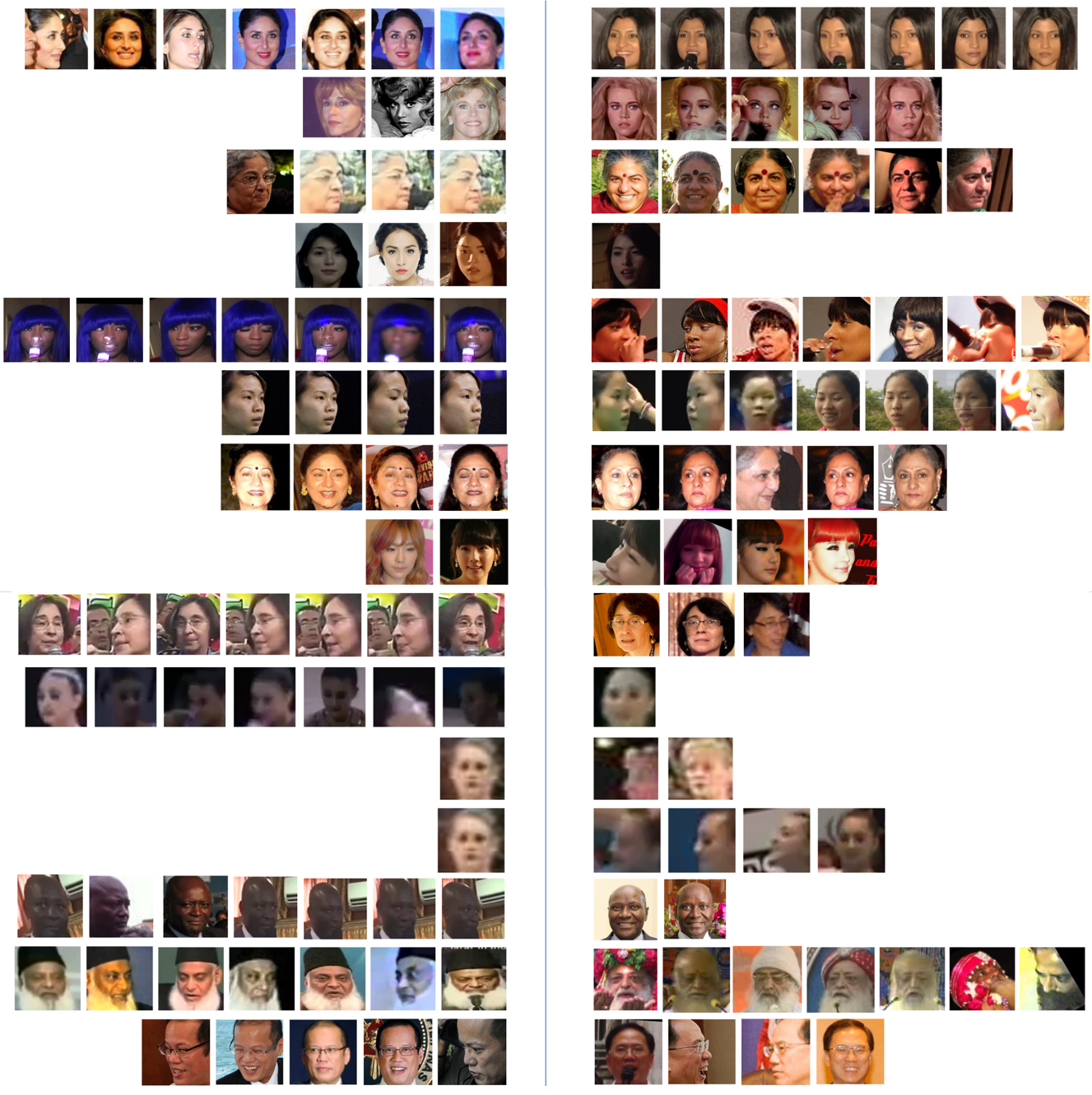}
\caption{Top $15$ False Positive verification pairs generated using \textbf{model-C} on IJB-C~\cite{mazeiarpa} dataset at false accept rate of $10^{-6}$}
\label{fig:bad_templates}
\end{figure*}

We present some observations about Crystal Loss, Quality Pooling and Quality Attenuation based on our experiments. Crystal Loss improves the performance over Softmax Loss by a large margin, since it specifically constrains the intra-class features to be close to each other in angular space. This is evident from Fig.~\ref{fig:score} which shows the distribution of similarity scores for verification pairs of IJB-C~\cite{mazeiarpa}. We observe that the match scores and non-match scores are more separated with Crystal Loss compared to Softmax Loss, which makes it valuable for face verification and identification. 

Quality Pooling and Quality Attenuation use face detection score to determine the authenticity of a given feature representation. The results show that the DCNN features are not invariant to the face size, pose or resolution. Similarity score from a high quality face pair is more reliable. 

We also visualize the top $15$ false positive verification pairs generated by \textbf{model-C} for IJB-C~\cite{mazeiarpa} dataset, that contributes to FAR of $10^{-6}$, in Fig.~\ref{fig:bad_templates}. We observe that most of the templates contain faces with ethnicity other than Caucasians. This shows that just like humans, the DCNN models suffer from the ``other-race effect''~\cite{o1994structural}, as most of the training data are biased towards Caucasian faces. One way to solve this problem is to incorporate sufficient amount of images from all the ethnicities in the training dataset.


\section{Conclusion}
\label{sec:conclusion}

In this paper, we proposed Crystal Loss that adds a simple, yet effective, $L_{2}$-constraint to the regular softmax loss for training a face verification system. The constraint enforces the features to lie on a hypersphere of a fixed radius characterized by parameter $\alpha$. We also provided bounds on the value of $\alpha$ for achieving a consistent performance. Additionally, we proposed Quality Pooling for generating better feature representation of a face video or template. Quality Attenuation also helps in improving the performance at very low FARs. Experiments on LFW, YTF and IJB datasets show that the proposed methods provide significant and consistent improvements and achieve state-of-the-art results on IJB-A~\cite{klare2015pushing}, IJB-B~\cite{whitelam2017iarpa} and IJB-C~\cite{mazeiarpa} datasets for both face verification and face identification tasks. In conclusion, Crystal loss is a valuable replacement for the existing softmax loss, for the task of face recognition. In the future, we would further explore the possibility of exploiting the geometric structure of the feature encoding using manifold-based metric learning.

\ifCLASSOPTIONcompsoc
  \section*{Acknowledgments}
\else
  \section*{Acknowledgment}
\fi

This research is based upon work supported by the Office of the Director of National Intelligence (ODNI), Intelligence Advanced Research Projects
Activity (IARPA), via IARPA R\&D Contract No. 2014-14071600012. The views and conclusions contained herein are those of the authors and should
not be interpreted as necessarily representing the official policies or endorsements, either expressed or implied, of the ODNI, IARPA, or the U.S. Government. The U.S. Government is authorized to reproduce and distribute reprints for Governmental purposes notwithstanding any copyright annotation
thereon.

\ifCLASSOPTIONcaptionsoff
  \newpage
\fi



%
{\small
\bibliographystyle{ieee}
\bibliography{nsfbib}
}

%

\begin{IEEEbiography}[{\includegraphics[width=1in,height=1.25in,clip,keepaspectratio]{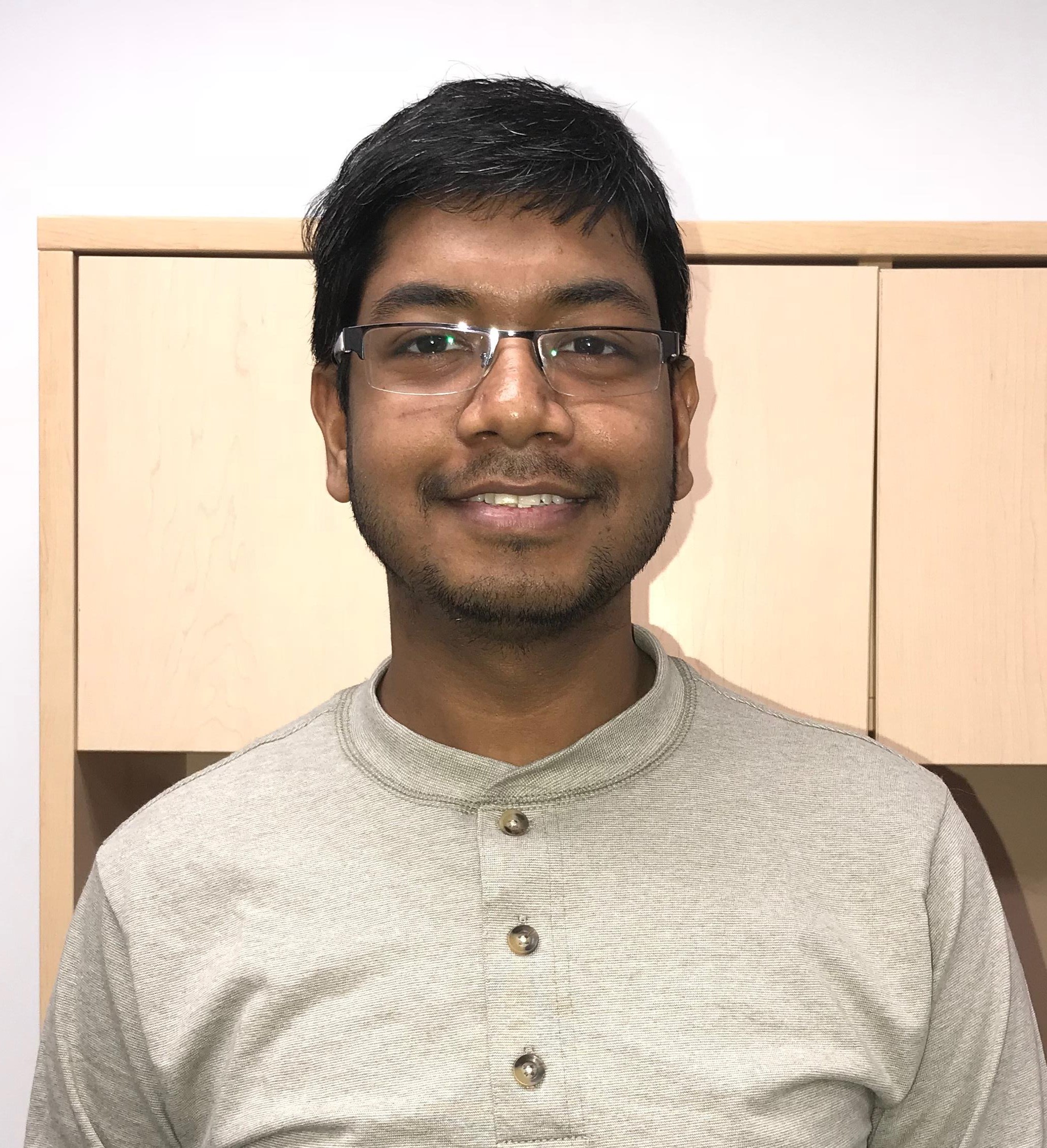}}]{Rajeev Ranjan}
received the B.Tech. degree in Electronics and Electrical Communication Engineering from Indian Institute of Technology Kharagpur, India, in 2012. He is currently a Research Assistant at University of Maryland College Park. His research interests include face detection, face recognition and machine learning. He received Best Poster Award at IEEE BTAS 2015. He is a recipient of UMD Outstanding Invention of the Year award 2015, in the area of Information Science. He received the 2016 Jimmy Lin Award for Invention.
\end{IEEEbiography}

\begin{IEEEbiographynophoto}{Ankan Bansal}
is a Ph.D. degree student at the University of Maryland, College Park. He received his B.Tech. and M.Tech. degrees in Electrical Engineering from the Indian Institute of Technology, Kanpur, in 2015. His research interests include multi-modal learning, action understanding, and face analysis.
\end{IEEEbiographynophoto}

\begin{IEEEbiography}[{\includegraphics[width=1in,height=1.25in,clip,keepaspectratio]{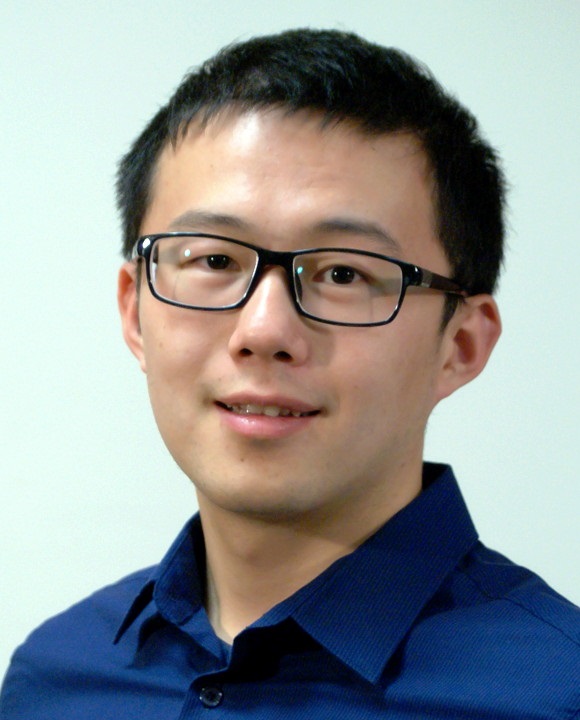}}]{Hongyu Xu}
received the BEng degree in Electrical Engineering from the University of Science and Technology of China in 2012 and the MS degree in Electrical and Computer Engineering from University of Maryland, College Park in 2016. He is currently a research assistant in the Institute for Advanced Computer Studies at the University of Maryland, College Park. His research interests include object detection, face recognition, object classification, and domain adaptation.
\end{IEEEbiography}

\begin{IEEEbiographynophoto}{Swami Sankaranarayanan}
is a Ph.D. candidate at University of Maryland College Park. He received his M.S degree from TU Delft, in 2012. His research interests include face analysis and adversarial machine learning.
\end{IEEEbiographynophoto}

\begin{IEEEbiographynophoto}{Jun-Cheng Chen}
is a postdoctoral research fellow at the University of Maryland Institute
for Advanced Computer Studies (UMIACS). He received the Ph.D. degree from University of Maryland, College Park in 2016. His current research interests include computer vision and machine
learning with applications to face recognition and facial analysis. He was a recipient of ACM Multimedia best
technical full paper award, 2006.
\end{IEEEbiographynophoto}

\begin{IEEEbiographynophoto}{Carlos D. Castillo}
is an assistant research scientist at the University of Maryland Institute
for Advanced Computer Studies (UMIACS). He received the Ph.D. degree from University of Maryland, College Park in 2012. His current research interests include stereo matching, multi-view
geometry, face detection, alignment and recognition.
\end{IEEEbiographynophoto}

\begin{IEEEbiography}[{\includegraphics[width=1in,height=1.25in,clip,keepaspectratio]{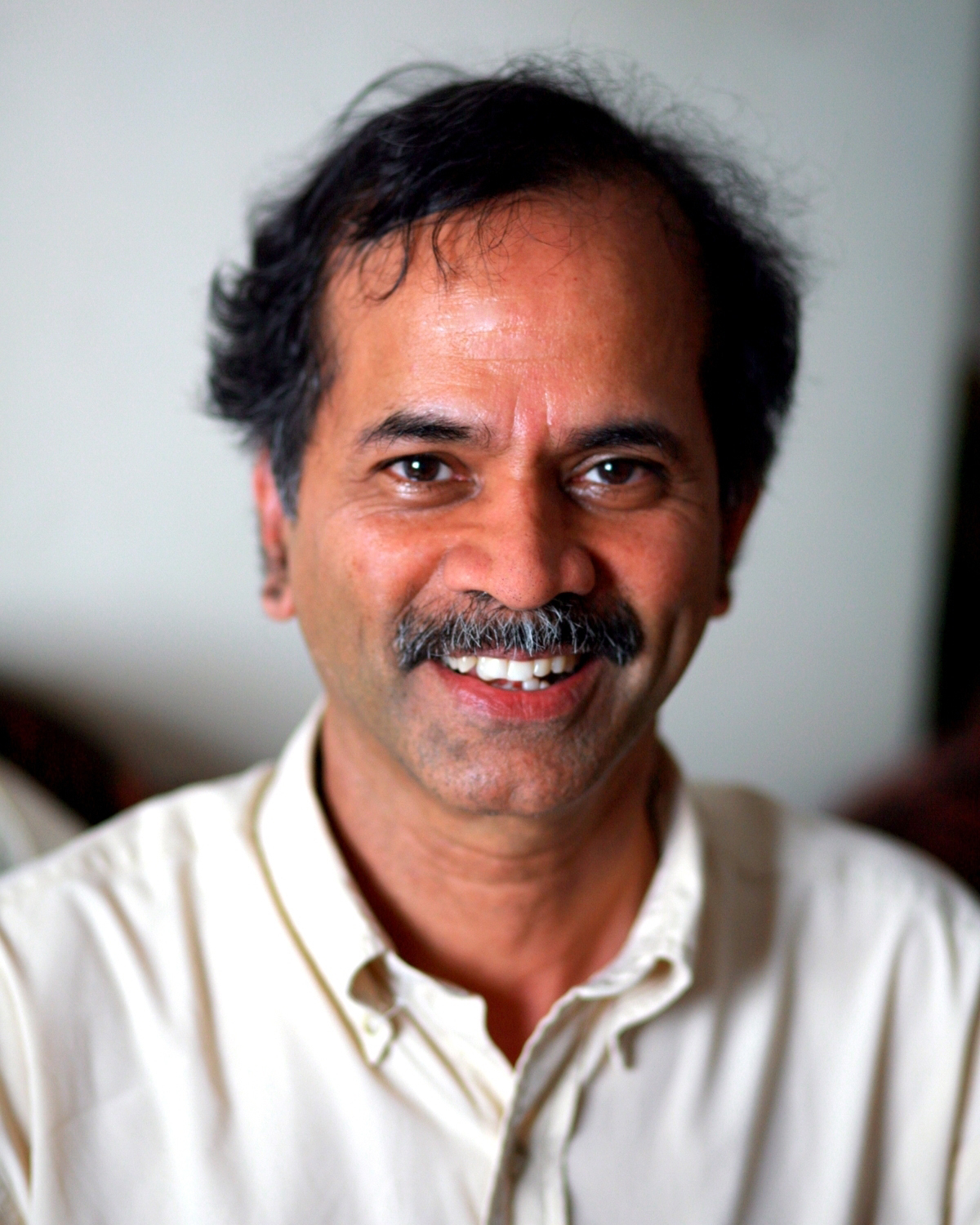}}]{Rama Chellappa}
is a Distinguished University Professor, a Minta Martin Professor of Engineering and Chair of the ECE department at the University of Maryland. Prof. Chellappa received the K.S. Fu Prize from the International Association of Pattern Recognition (IAPR). He is a recipient of the Society, Technical Achievement and Meritorious Service Awards from the IEEE Signal Processing Society and four IBM faculty Development Awards. He also received the Technical Achievement and Meritorious Service Awards from the IEEE Computer Society. At UMD, he received college and university level recognitions for research, teaching, innovation and mentoring of undergraduate students. In 2010, he was recognized as an Outstanding ECE by Purdue University. Prof. Chellappa served as the Editor-in-Chief of PAMI. He is a Golden Core Member of the IEEE Computer Society, served as a Distinguished Lecturer of the IEEE Signal Processing Society and as the President of IEEE Biometrics Council. He is a Fellow of IEEE, IAPR, OSA, AAAS, ACM and AAAI and holds four patents.
\end{IEEEbiography}






\end{document}